\documentclass[fleqn,10pt,letterpaper]{article}

\usepackage[left=2cm,right=2cm,top=2.25cm,bottom=2.25cm]{geometry}
\usepackage[utf8]{inputenc}
\usepackage{times}
\usepackage{mathptmx}
\usepackage{amsmath,amsfonts,amssymb}
\usepackage{graphicx,xcolor}
\usepackage{booktabs}
\usepackage[labelfont={bf,sf,small},labelsep=period,justification=raggedright]{caption}
\usepackage{enumitem}
\usepackage{microtype}
\definecolor{color1}{RGB}{0,0,0}
\definecolor{color2}{RGB}{0,20,20}

\usepackage[english]{babel}
\usepackage{csquotes}
\usepackage[style=apa,backend=biber]{biblatex}
\let\citeNP\nptextcite
\let\citeA\textcite
\let\cite\parencite
\usepackage{numprint}

\usepackage{pgfplots}
\pgfplotsset{compat=1.18}
\usetikzlibrary{patterns,shapes.geometric}

\usepackage{hyperref}
\usepackage[nameinlink]{cleveref}
\hypersetup{hidelinks,colorlinks,breaklinks=true,urlcolor=color2,citecolor=color1,linkcolor=color1,bookmarksopen=false,pdftitle={What Do Digital Traces Predict? Task Effects, Student Differences, and Critical Online Reasoning}}
\hypersetup{pdfauthor={Maxim Konca, Alexander Mehler}}

\newcommand{\code}[1]{\url{#1}}
\newcommand{\Rfinal}[1]{\textcolor{black}{#1}}

\usepackage{xspace}
\newcommand{\DIRQ}{\texttt{DIRQ}\xspace}
\newcommand{\TaSFC}{\texttt{TaSFC}\xspace}
\newcommand{\COH}{\texttt{COH}\xspace}

\title{What Do Digital Traces Predict? Task Effects, Student Differences, and Critical Online Reasoning}
\author{Maxim Konca\thanks{Corresponding author: konca@em.uni-frankfurt.de}\qquad Alexander Mehler\\[0.5ex]
  \normalsize Institute of Informatics and Mathematics, Goethe University, Frankfurt am Main, Germany}
\date{\normalsize Manuscript submitted for publication}

\begin{document}

\flushbottom

\maketitle

\begin{abstract}
Digital assessment traces are increasingly used to predict student performance, often evaluated by correlating predicted and observed scores pooled across tasks.
However, such correlations may reflect between-task differences rather than performance differences among students within tasks.
To investigate this, we analyzed up to 527 browsing sessions from 424 university students completing open-Internet Critical Online Reasoning tasks and assessed the cross-task stability of their ratings.
We evaluated the same out-of-fold predictions within and across tasks, compared three validation tiers up to leave-one-task-out, and tested incremental predictive contributions of consultation order and relational structure against matched permutation and randomization controls.
None of the twelve prespecified primary tests was significant: on held-out tasks, consulted-page features showed no significant association with relative student performance, nor did observed order or relational structure add predictive value.
Pooling observations across tasks substantially increased correlations, particularly for factual correctness, whose scores showed limited evidence of stable between-student differences across tasks.
However, sentence embeddings yielded modest positive correlations with generic response quality on held-out tasks, without establishing added predictive value from observed order or relational structure.
Our findings show that pooled predictive correlations may not reflect within-task student differences, highlighting the need for within-task evaluation and criterion generalizability audits.
\end{abstract}

\begin{quote}
\end{quote}

\section{Introduction}\label{sec:Introduction} %
Digital assessment environments (see \citeNP{Katz:2007}, for an early example; \citeNP{Sparks:et:al:2016}, and \citeNP{VanDamme:Zahner:2022}, for reviews; and \citeNP{Tenison:Sparks:2023}, for a recent study) capture both student outcomes, such as written responses, and process data, including search queries in internet-based information problem-solving tasks, accessed resources, and interaction sequences.
The competence assessed here is broadly critical thinking, understood as purposeful, self-regulatory judgment involving interpretation, analysis, evaluation, inference, and justification of the grounds for judgment \cite[p. 3]{Facione:1990}.
Applied to online information literacy, critical thinking becomes Critical Online Reasoning (COR): searching, selecting, evaluating, and reasoning from online information to reach justified conclusions \cite{Molerov:et:al:2020}.
Critical thinking, as addressed by COR research, is typically measured as a latent construct inferred from process observables (e.g., page visits, dwell times, scrolling, and page transitions) and product observables (e.g., linguistic features of written responses) collected during assessment tasks.
To infer COR from product variables, this approach uses rater scores of written responses as indicators of task-specific performance, which serves as an intermediary construct related to COR.
The problem is that a data-driven predictor may correlate highly with these scores by exploiting task-specific rather than task-independent information.
The challenge is therefore to demonstrate generalization first to unseen but related tasks and ultimately to unrelated tasks, while establishing construct validity.

This article addresses the first challenge by modeling COR-related task-specific performance across related tasks using both process and product observables.
In doing so, it revisits a longstanding problem in performance assessment research.
Generalizability Theory conceptualizes observed performance scores as arising from samples of tasks, occasions, raters, and measurement methods, with the intended scope of generalization determining which sources of variation constitute measurement error \cite{Cronbach:et:al:1972,Brennan:2001}.
In a study of elementary-level mathematics and science performance assessments, \citeA{Shavelson:et:al:1993} found that task-sampling variability contributed substantially more to measurement error than rater-sampling variability.
This finding highlights the importance of accounting for task-specific variation when predicting COR-related performance.
Although task-specific performance is informative, a single-task score need not reflect a stable, task-general student characteristic.
In particular, student-by-task interaction variance constitutes measurement error when generalizing student performance across tasks.

A parallel problem arises in predictive learning analytics.
Across nine university courses, \citeA{Gasevic:et:al:2016} found that predictive performance and predictor relevance varied between course-specific and generalized models, demonstrating that ignoring instructional contexts can distort estimated associations between learning-management-system features and academic success.
Similarly, the present study examines predictive generalization across assessment tasks using both process and product observables.
Specifically, it investigates whether these observables contain task-independent predictive information.

These two research traditions motivate a more stringent question than whether process or product observables correlate with observed rater scores.
Predictive modeling must distinguish task-context information from information that differentiates students performing the same task (within-task predictive validity).
It must also establish the incremental predictive value of each observable type and clarify what these observables represent.
Crucially, it must determine whether within-task predictive relationships generalize to unseen tasks (cross-task predictive generalizability).
More specifically, outcome scores derive from ratings of written responses, while both process- and product-related observables serve as predictors.
We audit the performance criteria and evaluate predictive generalization at three levels: observation, student, and task.
For process-related observables, we focus on consulted content pages, excluding search-results pages, which constitute approximately one-third of the captured pages; a sensitivity analysis includes the full browsing record.
Using written-response features as the baseline, we assess the incremental predictive value of consulted-page content, order-invariant features, consultation order, and graph-based page relations.
All models use matched observations for comparability.
Permutation and randomization controls, along with neural ablations, test whether temporal and relational structures add predictive information beyond page content.

In sum, our contribution is not to demonstrate task-sampling variability or limited predictive generalizability, both of which are already recognized challenges.
Rather, we connect these problems in COR-related performance assessment by testing whether models based on process observables and product observables differentiate students within tasks (within-task predictive performance) and generalize to unseen tasks (cross-task predictive generalizability), while examining how much pooled associations reflect task differences.
We further test whether structured process representations add predictive value beyond content representations ($\mathcal{C}$), using matched null representations as controls.
The study addresses the following research questions on within-task prediction and cross-task generalizability:
\begin{enumerate}
    \item[RQ1] \textbf{Product:} To what extent do linguistic features of written responses, not explicitly included in the rating criteria, predict COR-related performance as measured by rater scores?
    \item[RQ2] \textbf{Process content:} To what extent do linguistic features of consulted documents predict COR-related performance?
    \item[RQ3] \textbf{Process sequence:} Does consultation order add predictive information beyond the unordered set of consulted documents?
    \item[RQ4] \textbf{Process network:} Do graph-based representations of information-seeking trajectories add predictive information beyond document content and consultation order?
    \item[RQ5] \textbf{Generalizability:} To what extent do models evaluated under RQ1--RQ4 transfer to unseen tasks?
\end{enumerate}

\section{Related work}\label{sec:related}
\paragraph{The assessed construct.}
COR was conceptualized as a construct and assessment framework for university students \cite{Molerov:et:al:2020}.
Generic COR tasks were subsequently validated using several methods, including generalizability theory \cite{Nagel:et:al:2022,Nagel:2025}.
\citeA{Nagel:et:al:2022} found that person and person-by-task variance dominated task and rater effects, interpreting both as meaningful performance variation consistent with the assessment design.
Because variance components depend on the sampled students, tasks, raters, and scoring procedures, we re-estimate them here.
Student-by-task interaction variance may reflect meaningful task-specific performance differences but limits the interpretation of scores or predictions as indicators of task-general competence.
\citeA{Hartig:et:al:2026} document the rating procedure for generic COR tasks involving online information acquisition, evidence-based reasoning, and synthesis.
We use existing aggregate ratings and audit their underlying criteria rather than developing new ones.

\paragraph{Comprehension across multiple documents.}
Reading across multiple sources to answer questions is conceptualized as an interaction among reader, task, and texts rather than a reader attribute alone \cite{Rouet:et:al:2011,List:et:al:2019}.
The MD-TRACE model reflects this perspective and is particularly relevant to COR assessment because it treats task-model construction and evaluation of the resulting product as core processes \cite{Rouet:et:al:2011}.
COR-related performance scores are expected to be task dependent because tasks differ in their requirements, students' interpretations and responses, and the criteria used to evaluate performance.
The variance decomposition in \Cref{sec:results-criterion-audit} quantifies this variation without necessarily implying deficiencies in the rating rubric.

\paragraph{Process observables as performance predictors.}
Predicting performance from process observables is well established in learning analytics, with growing emphasis on interaction sequences rather than isolated events.
\citeA{Yu:et:al:2025} mined event sequences from 49 undergraduates completing a flipped-learning mathematics task, identifying six patterns correlated with post-test performance ($|r|\geq 0.2$), five involving multiple events and three negatively correlated.
However, these predictors were selected and evaluated on the same sample, without permutation controls.
Thus, the findings link sequence features to performance but do not establish the predictive contribution of event order itself.
Our study tests this contribution.
In the COR setting of \citeA{Molerov:et:al:2020}, \citeA{Schmidt:et:al:2020} collected eye-tracking and event-log data from 32 undergraduates completing a single COR task.
They used process mining to characterize information seeking and latent class analysis to identify two student profiles differing in process characteristics and rated task performance.
We likewise examine information seeking but focus on content-document visits and transitions, incorporating a content model absent from \citeA{Schmidt:et:al:2020}.

\paragraph{Aggregation and transfer across tasks.}
Pooled task correlations may differ from within-task correlations because they also reflect between-task variation.
Thus, a high pooled correlation may reflect within-task differences, between-task differences, or both, without reliably distinguishing students performing the same task.
This distinction also matters for prediction from process observables.
\citeA{Gasevic:et:al:2016} found that the same LMS feature could correlate positively or negatively with success depending on the course, and that significance differed between pooled and within-course models.
These findings show that pooled predictive performance does not necessarily identify features predictive within or across contexts.
A complementary methodological concern is the statistical reliability of predictive model comparisons.
Reviewing 87 predictive modeling experiments in MOOCs, \citeA{Gardner:et:al:2018} found that 52 lacked significance tests for model comparisons and none corrected for multiple testing.
A later review of 39 studies and 82 models of course-specific academic success found that performance was most often evaluated using accuracy alone  \cite{Arizmendi:et:al:2023}.
Neither review examines the distinction between within-task and between-task predictive associations central to this study.

The portability of predictive models across educational contexts remains a concern in learning analytics.
A review of 70 studies using log data from open-ended science and engineering environments identified task design, construct definition, feature extraction, and model validation as barriers to generalizability \cite{Wang:et:al:2023}.
More recently, \citeA{Xu:et:al:2026} trained models at one institution and tested them at another using standardized behavioral features from a shared online textbook, keeping the learning platform, content, and feature definitions relatively stable while varying the institutional context.
Our study examines complementary generalization: the assessment instrument and feature definitions remain fixed, while each of four tasks and its students are excluded from training in turn.
This design tests whether predictive performance persists when the evaluation task and its students are excluded from training.

\section{Materials and Methods}\label{sec:Methods}
\subsection{Data reconstruction and analysis populations}\label{sec:population}

We analyze student performance on four Critical Online Reasoning (COR) assessment tasks requiring students to consult online information and prepare written responses.
Each student-task observation comprises a written response and process observables recorded during task completion.
The four tasks, drawn from \citeA{Hartig:et:al:2026}, examine health risks from cell-phone towers (FM), mosquito-bite treatment with heat sticks (HS), Tetra Pak recyclability (TP), and poisoning from a traditional German dish (GS).
Page visits were reconstructed from page-opening logs and assigned to individual students and tasks.

We extracted linguistic features from process and product texts using Coh-Metrix \cite{Graesser:et:al:2004,McNamara:et:al:2014}.
We chose Coh-Metrix for its established use in linguistic analysis, cross-linguistic applicability, and standardized feature set, avoiding an exhaustive comparison of alternative models.
As the tasks and most texts were in German, we \Rfinal{computed the Coh-Metrix features using German-language text processing}, excluding non-German texts and those shorter than 50 words under an inherited minimum-content rule before constructing either feature representation.
\Rfinal{Reconstruction rules, URL matching, classifiers, item mapping, model settings, and ablation designs are documented in the Supplement.}

\paragraph{Page sets.}\label{sec:universes}
Visited pages were classified by URL as either search engine results pages (SERPs) or content pages, the latter being a pragmatic rather than a comprehensive genre category.
Page set $\mathsf{A}=\mathsf{C}\cup\mathsf{S}$ comprises all pages retained after preprocessing, partitioned into content pages ($\mathsf{C}$) and SERPs ($\mathsf{S}$), with $\mathsf{C}\cap\mathsf{S}=\emptyset$.
Page sets determine which pages are considered, whereas analysis populations comprise the student-task observations eligible for analysis within each page set.\footnote{%
We use \enquote{page set} to avoid confusion with the universe of admissible observations in generalizability theory \cite{Cronbach:et:al:1972,Brennan:2001}.}
A student-task observation is eligible for a given page set if at least two pages meet the feature-extraction requirements.
As RQ2--RQ4 focus on consulted documents, $\mathsf{C}$ was designated the primary page set if two prespecified criteria were met: at least 30 eligible observations per task and at most 25\% two-page trajectories.
\Cref{sec:results-sample} reports whether these criteria were met.

\paragraph{Ratings and annotations.}
The study uses observed rater scores assigned by trained student raters using the generic-COR rating procedure \cite{Hartig:et:al:2026}.
The primary analyses use the supplied criterion-level aggregate scores; neither the rating procedure nor the original aggregation was modified.
Topic, genre, and emotion labels were assigned during data preparation using the classifiers documented in the supplement.

\subsection{Performance outcomes and criterion audit}\label{sec:targets}
\Rfinal{The four tasks share a rating rubric with generic criteria worded identically across tasks and task-specific factual-correctness criteria.
After rescaling the selected criteria to $[0,1]$, we formed three unweighted composites: \textit{Domain-Independent Response Quality} (\DIRQ; general linguistic and formal quality), the mean of response fulfillment, source referencing, and \textit{COHerence} (\COH; semantic connectedness and conceptual consistency of the response); \textit{Task-Specific Factual Correctness} (\TaSFC; factual correctness relative to task requirements), the mean of the task's factual-correctness criteria; and \textit{Overall}, the mean of all selected criteria.
Reliability and agreement coefficients describe the ratings but do not weight the composites.
The primary outcomes are \DIRQ, \TaSFC, and \COH.
\COH, the only single criterion among them, is analyzed separately although it enters \DIRQ, so the two analyses are not independent.
Response fulfillment and source referencing are analyzed as exploratory components of \DIRQ.}

Secondary outcomes include the unstandardized Overall composite and within-task standardized \DIRQ, \TaSFC, and Overall scores.
Standardized scores are defined as $z_{it}=(x_{it}-\mu_t)/\sigma_t$, with mean $\mu_t$ and standard deviation $\sigma_t$ computed from all valid ratings for task $t$, independently of the modeling cohort.
Retrospective standardization equalizes task means and standard deviations but does not ensure equivalent score interpretations across tasks.

The rating-only audit uses the 644 of 664 rated responses that have valid \DIRQ, \TaSFC, and \COH scores to assess cross-task rank-order consistency and estimate variance components attributable to students, student-by-task interactions, raters, and rater-by-task interactions.
Of the 144 students observed on two tasks, 136 had valid scores for both.

Each variance component is tested using a likelihood-ratio test comparing the fitted model with a reduced model omitting that component.
Projected dependability coefficients are reported without confidence intervals and interpreted solely as guidance for assessment design.

\subsection{Predictor feature representations for RQ1--RQ4}\label{sec:representations}
Visited pages (process-related observables) and written responses (product-related) are each represented using two feature sources.
The primary feature source provides 17 cohesion, lexical, and syntactic indices derived from a German adaptation of Coh-Metrix \cite{Graesser:et:al:2004}, expressed as rates, incidences, or means rather than raw counts.
Features were screened for missingness, variance, and length sensitivity independently of rater scores and fixed before predictive modeling.
Screening excluded word, sentence, and paragraph counts and mean sentence and paragraph lengths, retaining only mean word length in letters as a descriptive index.
The second feature source uses 384-dimensional sentence embeddings from \code{paraphrase-multilingual-MiniLM-L12-v2} \cite{Reimers:et:al:2019,Reimers:et:al:2020} for sensitivity analysis.
All embeddings were retrieved from a single frozen, content-addressed cache.
The indices are listed in \Cref{tab:coh-indices}.
For student-task observation $i$ and page set $\mathsf{U}\in\{\mathsf{A},\mathsf{C},\mathsf{S}\}$, let $n_i^\mathsf{U}$ denote the number of associated pages and $\mathbf{x}_{ij}^\mathsf{U}$ the feature vector of the $j$-th page.
The unordered page representation of $i$ for $\mathsf{U}$ is the element-wise mean of these vectors and is invariant to permutations of page order:
\begin{equation}
\mathbf{c}_i^\mathsf{U} = \frac{1}{n_{i}^\mathsf{U}}\sum_{j=1}^{n_{i}^\mathsf{U}} \mathbf{x}_{ij}^\mathsf{U}.
\end{equation}
Primary analyses use the content-page set $\mathsf{U} \gets \ \mathsf{C}$; sensitivity analyses use the all-page set $\mathsf{U} \gets \ \mathsf{A}$.

To represent process-related observables beyond page content, $\mathbf{c}_i^\mathsf{U}$ is concatenated with up to 34 order-invariant visit descriptors $\mathbf{v}_i^\mathsf{U}$, yielding $[\mathbf{c}_i^\mathsf{U},\mathbf{v}_i^\mathsf{U}]$.
These include log-transformed visit and unique-URL counts, revisit rates, and visit proportions by page topic, genre, and emotion, with missing labels treated as separate categories.
Preprocessing parameters are fitted only on training folds, while visit descriptors are computed separately for each page set.
Including $\mathbf{v}_i^\mathsf{U}$ in the baseline limits the attribution of order-sensitive and relational gains to differences in session size or annotation composition.

RQ2 focuses on $\mathbf{c}_i^\mathsf{U}$ without $\mathbf{v}_i^\mathsf{U}$ to isolate the predictive contribution of consulted page content from additional visit descriptors.
An exploratory comparison of $[\mathbf{c}_i^\mathsf{U},\mathbf{v}_i^\mathsf{U}]$ with $\mathbf{c}_i^\mathsf{U}$ assesses whether visit descriptors provide incremental predictive information beyond aggregated page content.

For RQ3, visit order is encoded by 16 features $\mathbf{o}_i^\mathsf{U}$ capturing trajectory positions, displacement, efficiency, and directional consistency in the space of the first four principal components of the page vectors.
The control $\tilde{\mathbf{o}}_i^{\mathsf{U}}$ is obtained by averaging these features over 32 random permutations of the visit sequence, while keeping $\mathbf{c}_i^\mathsf{U}$ and $\mathbf{v}_i^\mathsf{U}$ fixed.

For RQ4, page relations are encoded by 22 features $\mathbf{g}_i^\mathsf{U}$ capturing annotation agreement, cosine similarity between page pairs, weighted node degrees, and two normalized graph-Laplacian measures.
The control $\tilde{\mathbf{g}}_i^{\mathsf{U}}$ is obtained by averaging the relation features over 32 random joint permutations of the annotation labels and similarity matrix, while keeping the page vectors, temporal adjacency, and observed visit order fixed.
These permutations preserve annotation-label and pairwise-similarity distributions while disrupting their alignment with the observed consultation path.

In sum, for student-task observation $i$ and page set $\mathsf{U}$, we distinguish four types of feature vectors: content-related ($\mathbf{c}_i^\mathsf{U}$), order-invariant visit-related ($\mathbf{v}_i^\mathsf{U}$), order-sensitive visit-related ($\mathbf{o}_i^\mathsf{U}$), and relational ($\mathbf{g}_i^\mathsf{U}$).
We denote the corresponding representation levels by $\mathcal{C}$, $\mathcal{V}$, $\mathcal{O}$, and $\mathcal{G}$, respectively, and introduce $\mathcal{W}$ for the representation of students' written responses.

\subsection{Estimators, validation tiers, and scoring}\label{sec:estimators}\label{sec:tiers}
For fixed-length features, we use random forest regression \cite{Breiman:2001} as the primary estimator, with radial basis function support vector regression and partial least squares regression for sensitivity analyses, all with fixed hyperparameters and without tuning.
\Rfinal{%
As exploratory RQ3 and RQ4 benchmarks, we fit a single-layer gated recurrent unit (GRU) \cite{Cho:et:al:2014} and a message-passing graph network \cite{Gilmer:et:al:2017}, with matched order and relation ablations.}

We fit neural models to predict \Rfinal{\DIRQ, \TaSFC, and \COH from sequences or graphs of page vectors $\mathbf{x}_{ij}^\mathsf{U}$}, using three random seeds.
All predictor transformations, including imputation, scaling, principal components, annotation levels, and representation adapters, are fitted only on the respective training partition.
\Rfinal{Task standardization transforms targets, not predictors, and its constants include evaluation data, so standardized outcomes are retrospective targets unavailable in prospective prediction.}

We use three evaluation tiers (T1--T3) with increasingly restrictive train-test separation and fixed split manifests based on page set $\mathsf{A}$, shared across outcomes, representations, estimators, and feature sources.
\begin{description}
    \item[Tier T1, observation-level.] %
    We partition student-task observations into five task-stratified folds and repeat this procedure five times.
    Observations from the same students and tasks may occur in both training and test sets.
    \item[Tier T2, student-grouped.] %
    We use five-fold, \Rfinal{task-stratified,} student-grouped cross-validation with five repetitions, ensuring student-disjoint train-test splits while retaining all tasks in training.
    \item[Tier T3, task holdout.] %
    We hold out each task in turn, excluding all observations from its students from training, thereby evaluating transfer to an unseen task with student-disjoint splits.
\end{description}

T2 prevents student overlap between training and test sets, following blocked cross-validation principles for hierarchical and repeated-measures data \cite{Roberts:et:al:2017}.
T3 retains student-level separation while training on three tasks and testing exclusively on the fourth.
Treating each task as a distinct domain, the training and test sets are domain-disjoint, corresponding to the domain-generalization setting described by \citeA{Koh:et:al:2021}.
As all four tasks belong to the same study, T3 provides retrospective leave-one-task-out transfer evaluation \Rfinal{(RQ5)} rather than prospective or independent external validation.

Across all tiers, our primary metric is the within-task Spearman rank correlation ($\rho$) between predicted and observed scores.
For page set $\mathsf{U}$, let $X^{\mathsf{U}}=(f(\mathbf{x}_i^{\mathsf{U}}))_{i\in I}$ denote the potentially transformed feature vectors for observations indexed by $I$, where $f$ is learned during training when required by the estimator and otherwise equals $\mathrm{id}$.
We denote the within-task Spearman correlation between observed scores and model predictions by $\rho(X^{\mathsf{U}})$, so $\rho(X^{\mathsf{U}})-\rho(Y^{\mathsf{W}})$ compares predictive performance between representations $X$ and $Y$ for page sets $\mathsf{U},\mathsf{W}\in\{\mathsf{A},\mathsf{C},\mathsf{S}\}$.
Out-of-fold predictions are scored separately within each task.
The four task-specific Spearman correlations are weighted equally and subsequently averaged across repeats and, for neural models, random seeds.
Each task must have at least 30 valid ratings.
For %
T1 and T2, Spearman correlation is also computed on predictions pooled across tasks.
Comparing pooled and within-task correlations indicates whether apparent predictive performance may partly reflect between-task score differences rather than the ability to rank students within tasks.

\subsection{RQ-specific contrasts and statistical inference}\label{sec:rq-design}\label{sec:inference}
Evaluating the predictive contribution of structural information requires matched non-structural input features and data splits, plus structure-agnostic baselines \cite{Errica:et:al:2020}.
Accordingly, each contrast in our study holds the estimator or architecture, observations, and data splits fixed, varying only the representational component under evaluation.

The primary RQ2--RQ4 analyses use page set $\mathsf{C}$ \Rfinal{and the same eligible} student-task observations.
For RQ2, we compare the content representation $\mathcal{C}^\mathsf{C}$, the written-response representation $\mathcal{W}$, and their concatenation $\mathcal{C}^\mathsf{C} + \mathcal{W}$, where %
$+$ denotes feature concatenation rather than vector addition.
For student-task observation $i$, the %
feature vectors are $\mathbf{c}_i^\mathsf{C}$, $\mathbf{w}_i$, and $[\mathbf{c}_i^\mathsf{C},\mathbf{w}_i]$, respectively.
The fixed-length representations for RQ3 and RQ4 cumulatively combine feature components:
\begin{equation}
\mathcal{C}^\mathsf{C}
\longrightarrow \mathcal{C}^\mathsf{C} + \mathcal{V}^\mathsf{C}
\longrightarrow \mathcal{C}^\mathsf{C} + \mathcal{V}^\mathsf{C} + \mathcal{O}^\mathsf{C}
\longrightarrow \mathcal{C}^\mathsf{C} + \mathcal{V}^\mathsf{C} + \mathcal{O}^\mathsf{C} + \mathcal{G}^\mathsf{C} \ .
\end{equation}
The exploratory increment from $\mathcal{C}^\mathsf{C}$ to $\mathcal{C}^\mathsf{C} + \mathcal{V}^\mathsf{C}$ establishes the order-invariant baseline for structural comparisons.
The next two increments add observed visit order (RQ3) and relational information (RQ4), respectively.
Each contrast uses the same eligible observations for both compared specifications.

\paragraph{RQ1--RQ2: document and response content.}\label{sec:rq1}
\Rfinal{RQ2 compares consulted pages with written responses; RQ1 is addressed descriptively and by secondary contrasts.}
The primary contrasts assess consulted-page predictive performance, $\rho(\mathcal{C}^\mathsf{C})$, and its increment beyond written-response features, $\rho(\mathcal{C}^\mathsf{C}+\mathcal{W})-\rho(\mathcal{W})$.
Positive increments indicate improvement; negative increments indicate deterioration.
Secondary contrasts assess the written-response increment, $\rho(\mathcal{C}^\mathsf{C}+\mathcal{W})-\rho(\mathcal{C}^\mathsf{C})$, and the direct performance difference, $\rho(\mathcal{C}^\mathsf{C})-\rho(\mathcal{W})$.
Because written-response features and ratings derive from the same text, their association may partly reflect shared-source overlap.

\paragraph{RQ3: consultation order.}\label{sec:rq2}
We denote the permutation-control representation level by $\widetilde{\mathcal{O}}^\mathsf{U}$ and its feature vector by $\tilde{\mathbf{o}}_i^\mathsf{U}$.
We compare $\mathcal{C}^\mathsf{C}+\mathcal{V}^\mathsf{C}+\mathcal{O}^\mathsf{C}$ against the order-invariant baseline $\mathcal{C}^\mathsf{C}+\mathcal{V}^\mathsf{C}$ and permutation control $\mathcal{C}^\mathsf{C}+\mathcal{V}^\mathsf{C}+\widetilde{\mathcal{O}}^\mathsf{C}$.
Evidence that observed consultation order adds predictive value requires positive differences in both comparisons.
We also use exploratory matched neural ablations to assess consultation-order effects in graph and sequence models.
Comparisons across fixed-length, graph, and GRU models are descriptive benchmarks because architectures and input representations differ.

\paragraph{RQ4: relational organization.}\label{sec:rq3}
We denote the permutation-control representation level by $\widetilde{\mathcal{G}}^\mathsf{U}$ and its %
vector by $\tilde{\mathbf{g}}_i^\mathsf{U}$.
We compare $\mathcal{C}^\mathsf{C}+\mathcal{V}^\mathsf{C}+\mathcal{O}^\mathsf{C}+\mathcal{G}^\mathsf{C}$ with %
the order-sensitive baseline $\mathcal{C}^\mathsf{C}+\mathcal{V}^\mathsf{C}+\mathcal{O}^\mathsf{C}$ and the permutation-control specification $\mathcal{C}^\mathsf{C}+\mathcal{V}^\mathsf{C}+\mathcal{O}^\mathsf{C}+\widetilde{\mathcal{G}}^\mathsf{C}$.
Evidence that relational organization adds predictive value requires positive differences in both comparisons.
Exploratory matched graph ablations compare relational, temporal, and randomized-relation graphs \Rfinal{with the same message-passing network}.

\paragraph{Inference.}
We define a specification contrast as the difference in mean within-task Spearman correlations between predictor specifications evaluated on identical held-out observations.
\Rfinal{Confidence intervals use a student-cluster percentile bootstrap \cite{Efron:et:al:1993}; contrasts are tested by paired permutations within students, and individual correlations by permuting ratings within tasks.}

For RQ3 and RQ4, we use conjunction tests (logical \textsc{and} tests): the null hypothesis states that at least one of the two component effects is non-positive, whereas the alternative requires both to be positive.
The first component tests incremental predictive value relative to the preceding baseline.
The second tests whether the observed representation outperforms its matched permutation or randomization control, distinguishing structural contributions from gains due merely to additional features.
Logged events do not explain the processes generating them, nor does predictive value establish their interpretation.
We therefore do not interpret order and relational features as specific cognitive strategies but test whether observed structural information adds predictive value beyond matched controls.

\Rfinal{The conjunction $p$-value is the larger component $p$-value if both estimated effects are positive, and 1 otherwise \cite{Berger:1982}.}
RQ2 uses separate tests because its two quantities address distinct estimands: whether consulted-page content predicts performance and whether it adds predictive information beyond the written response.
Neither claim is conditional on the other.

We use paired student-cluster bootstraps to assess differences between evaluation tiers and the \emph{pooling gap}, defined as the difference between pooled and within-task Spearman correlations.
A separate family of twelve one-sided tests assesses whether predictive performance in T1 or T2 exceeds that in T3 by more than $0.05$.
These tests cover the consulted-page representation $\mathcal{C}^\mathsf{C}$ and the order-invariant baseline $\mathcal{C}^\mathsf{C}+\mathcal{V}^\mathsf{C}$.

\paragraph{Prespecified primary families.}
The prespecified primary tests use page set $\mathsf{C}$, Tier~3, Coh-Metrix features, random forests, and three outcomes: \DIRQ, \TaSFC, and \COH.
RQ2 comprises six tests: $\rho(\mathcal{C}^\mathsf{C})>0$ and $\rho(\mathcal{C}^\mathsf{C}+\mathcal{W})-\rho(\mathcal{W})>0$, each evaluated for all three outcomes.
RQ3 and RQ4 each comprise three outcome-specific conjunction tests.
We apply Holm correction \cite{Holm:1979} separately within each research-question family, using the conjunction $p$-values rather than their individual component $p$-values.
Exploratory tests are corrected for multiple comparisons within their %
prespecified groups and page sets.
We also report source-wide Benjamini--Hochberg-adjusted $p$-values descriptively, not for confirmatory inference.
We define narrow hypothesis families matching research questions and prespecified contrasts.
The confirmatory analysis includes twelve tests in three families; the largest family overall contains 47 tests.

\section{Results}\label{sec:Results}

\subsection{Analysis sample and process-data coverage}\label{sec:results-sample}

\begin{figure}[t]
    \centering
    \includegraphics[width=1\textwidth]{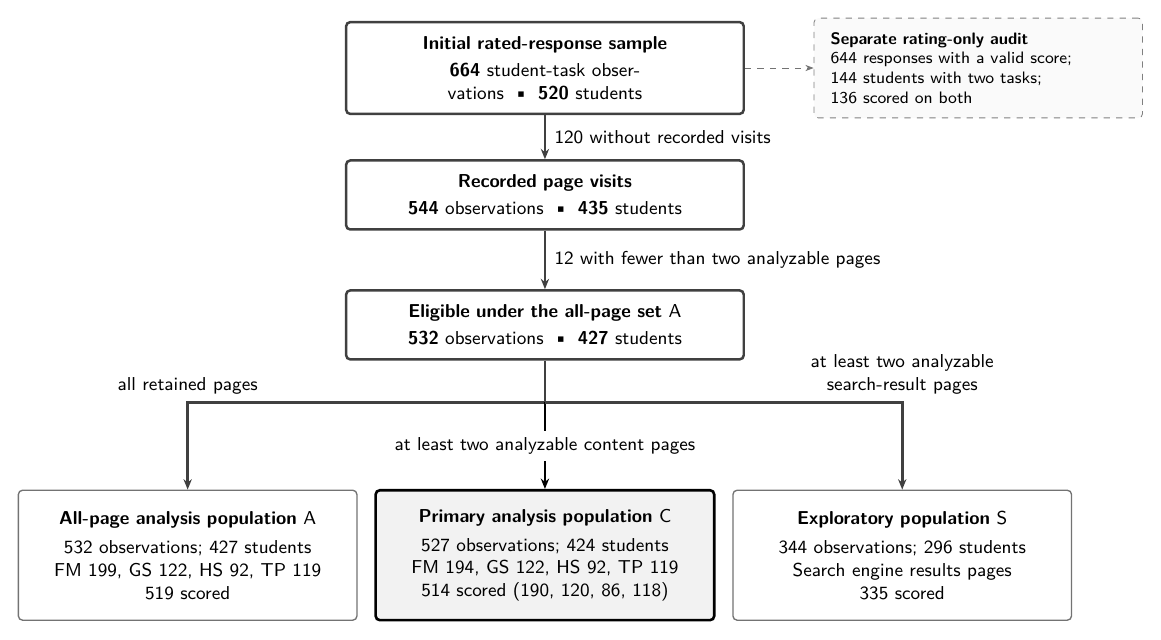}
    \caption{Sample flow. The rating-only audit includes all validly rated responses. Eligibility requires at least two analyzable pages from the respective page set.
    $\mathsf{A}=\mathsf{C}\cup\mathsf{S}$, $\mathsf{C}\cap\mathsf{S}=\emptyset$, partitions all preprocessed retained pages into content pages ($\mathsf{C}$) and SERPs ($\mathsf{S}$).
    Scored observations include all three primary outcomes and enter model fitting and evaluation.}
    \label{fig:sample-flow}
\end{figure}

\Cref{fig:sample-flow} shows the selection of student-task observations.
Of 664 rated responses, 544 from 435 students had at least one recorded page visit; 532 had at least two analyzable pages.
Page set $\mathsf{C}$ yields 527 eligible observations from 424 students at three German universities; 514 have valid scores on all three primary outcomes, with at least 86 scored observations per task.
The prespecified criteria for $\mathsf{C}$ were met: the smallest task had 92 eligible observations, and 474 of 527 trajectories (90\%) contained at least three pages.
Two-page trajectories, with structurally zero relational contrast, comprised 10.1\% of eligible observations in $\mathsf{C}$ and 1.9\% in $\mathsf{A}$.

\Rfinal{Of the sources students named in their responses, 76\% matched recorded visits with analyzable text, whereas 15\% came from sessions without recorded visits.}

\subsection{Criterion audit: rating variability and cross-task consistency}\label{sec:results-criterion-audit}
Before evaluating the models, we assess between-student variation in criterion ratings and the stability of relative performance across tasks.
This rating-only audit (see \Cref{sec:targets}) preceded the frozen validation run.

\begin{figure}[t]
    \centering
    \includegraphics[width=0.8\linewidth]{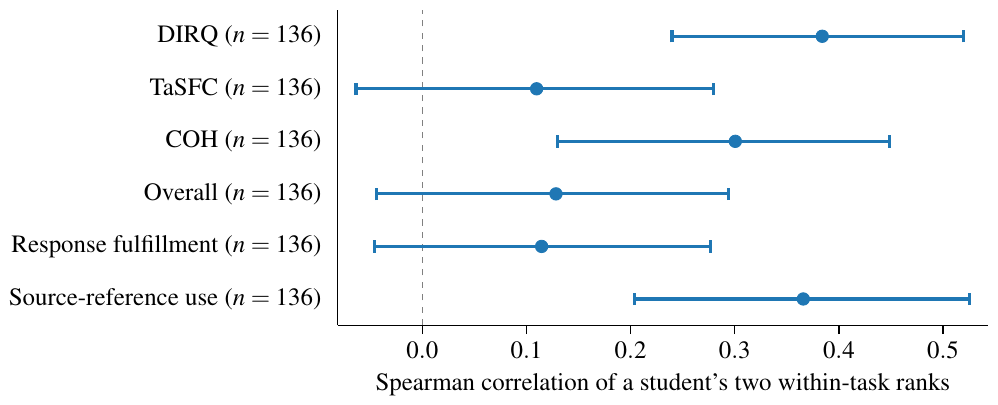}
    \caption{Cross-task rank-order consistency of rating outcomes and criteria.
    Points show Spearman correlations between task-specific score ranks for 136 students with valid scores on both tasks (of 144 completing two tasks).
    Ranks are based on all valid responses per task; error bars show 95\% student-level bootstrap confidence intervals.
    Task pairs are ordered alphabetically, not chronologically.
    \DIRQ, \TaSFC, and \COH are primary outcomes; Overall, response fulfillment, and source-reference use are additional measures.}
    \label{fig:criterion-cross-task}
\end{figure}

Cross-task rank-order consistency (the extent to which students maintained their relative standing across the two tasks they completed) varied substantially across the three primary outcomes (\Cref{fig:criterion-cross-task}).
Of 520 students with rated responses, 144 completed two tasks, yielding 136 student-level pairs with valid scores for cross-task analysis.
We apply Holm correction to the three outcome-specific $p$-values.
Significant positive cross-task correlations were observed for \DIRQ ($\rho=0.38$, 95\% CI $[0.24,0.51]$, $p_{\mathrm{Holm}}<0.001$) and \COH ($\rho=0.30$, 95\% CI $[0.13,0.44]$, $p_{\mathrm{Holm}}=0.001$).
By contrast, the cross-task correlation for \TaSFC was smaller and not statistically significant ($\rho=0.11$, 95\% CI $[-0.07,0.28]$, $p_{\mathrm{Holm}}=0.204$).
Across the six task pairs, correlations for \DIRQ and \COH were consistently positive, although those for \DIRQ ranged from $+0.07$ to $+0.57$.
In contrast, \TaSFC correlations were negative for two task pairs.
With only 17--27 students per task pair, these imprecise correlations illustrate cross-task heterogeneity rather than establish pair-specific consistency.
Failure to reject the null hypothesis does not establish that the association is zero.
The 95\% confidence interval ($[-0.07,0.28]$) includes both zero and $0.25$, \Rfinal{a value} within the range estimated for \COH.
Thus, the data do not distinguish the absence of cross-task consistency from a moderately positive association.
The evidence for \TaSFC's cross-task consistency is therefore inconclusive.
Stronger evidence of \TaSFC's lower cross-task consistency comes from the variance decomposition below: at least half the variance reflects student-by-task interactions, while stable between-student variance is not significantly different from zero.

\begin{table}[t]
    \centering
    \caption{%
    \Rfinal{Restricted-maximum-likelihood} variance components for the three primary outcomes and two additional criteria (response fulfillment and source-reference use), expressed as percentages of total variance.
    Brackets show post hoc 95\% profile-likelihood confidence intervals for non-residual components.
    The rater component is particularly uncertain: six raters contributed, four providing 88\% of ratings.
    Estimates sum to approximately 100\%, subject to rounding.}
    \label{tab:criterion-variance}
    \small
    \begin{tabular}{lrrrrr}
        \toprule
        Outcome & Student & Student $\times$ task & Rater & Rater $\times$ task & Residual \\
        \midrule
        \DIRQ                     & 23 [14, 29] & 23 [17, 30] & 9 [0, 60] & 8 [3, 29] & 37 \\
        \TaSFC                    & 8 [0, 20] & 64 [53, 69] & 2 [0, 16] & 5 [2, 14] & 21 \\
        \COH             & 19 [11, 24] & 14 [8, 21] & 19 [6, 87] & 3 [1, 10] & 45 \\
        Response fulfillment  & 6 [0, 12] & 20 [14, 27] & 0 [0, 11] & 15 [8, 35] & 59 \\
        Source-reference use  & 36 [23, 44] & 43 [34, 51] & 1 [0, 21] & \Rfinal{$<$1} [0, 2] & 20 \\
        \bottomrule
    \end{tabular}
\end{table}

The variance decomposition helps explain these differences (\Cref{tab:criterion-variance}).
\Rfinal{%
\Rfinal{With only 136 students scored on two tasks,} separating stable student from student-by-task variance is particularly uncertain.}

Despite the uncertainty, \TaSFC shows a pronounced difference between its student-by-task interaction variance (64\%) and student variance (8\%), suggesting that \TaSFC scores depend substantially more on the particular student-task combination than on stable between-student differences.
By contrast, \DIRQ and \COH have similar estimated student variance shares (23\% and 19\%, respectively).
Because each student completed each task only once, the student-by-task interaction component confounds task-specific and occasion-specific variation.
This student-by-task/occasion variation indicates limited cross-task consistency in \TaSFC, without distinguishing task-sampling from occasion effects.
Stable between-student variation is established for \DIRQ and \COH, potentially supporting cross-task transfer, but not for \TaSFC.
However, the stable components cannot be ranked across outcomes: \TaSFC's interval reaches 20\%, near \DIRQ's estimate of 23\%.

\Rfinal{Rater components were the least precise estimates; among the primary outcomes, only \COH showed a significant rater main effect (19\%, $p=0.001$).}

A COR validation study likewise found substantial person-by-task variance and small rater effects \cite{Nagel:et:al:2022}, while task-sampling effects are documented in other performance assessments \cite{Shavelson:et:al:1993}.
Our findings agree for \TaSFC and \DIRQ, whose rater main effects are not significant, but differ for \COH, whose effect is significant.
However, the limited number of raters \Rfinal{makes} these estimates highly uncertain.
The implication depends on the intended interpretation.
Student-by-task variation contributes to task-specific performance but limits its interpretation as a task-general student attribute.
For \TaSFC, stable between-student variance is estimated at 8\% (95\% CI [0\%, 20\%]), leaving a potentially substantial task-general component unconfirmed.
\Rfinal{%
\Rfinal{Single-task generalizability coefficients with three raters were} $0.39$ for \DIRQ, $0.40$ for \COH, and $0.10$ for \TaSFC \cite{Cronbach:et:al:1972}.
Four tasks would raise \DIRQ and \COH to $0.70$, whereas \TaSFC would require about 22; these projections assume a stable between-student component, established only for \DIRQ and \COH.}

\Rfinal{%
A post-hoc analysis showed that response length correlated within tasks with \COH ($\rho=0.34$), source-reference use ($0.27$), \Rfinal{response fulfillment ($0.11$),} and \DIRQ ($0.39$), but barely with \Rfinal{the 32 factual-correctness criteria} (median $0.03$).
Adjusting for 17 response-level Coh-Metrix indices left correlations with \COH and \DIRQ unchanged ($0.337$ and $0.393$), but competence and rater sensitivity to length cannot be disentangled.}

Two implications follow for the predictive analyses.
First, Tier~3 provides a meaningful test of whether models trained on other tasks can predict \DIRQ and \COH on a held-out task, for which stable between-student variation is established.
Second, \TaSFC's stable between-student component is not distinguishable from zero, limiting its task-general interpretation and complicating Tier~3 evaluation.
Nevertheless, held-out task traces may still support prediction of task-specific \TaSFC.
Weak correlations may thus reflect limitations of the predictors, the outcome criterion, or both.

\subsection{Prespecified primary families}\label{sec:rq-results}
None of the twelve prespecified primary tests survived Holm correction (\Cref{tab:primary-results}); the smallest adjusted $p$-value was $0.803$.
These tests used Tier~3, Coh-Metrix features, random forests, and content-page set $\mathsf{C}$.
However, results outside this setting were not uniformly near zero: at Tier~3, sentence embeddings of the same pages showed a positive association with \DIRQ (\Cref{sec:source-results}).

\begin{table}[t]
  \centering
  \caption{%
  Twelve prespecified primary tests for page set $\mathsf{C}$ at Tier~3, using Coh-Metrix features and random forests (527 eligible observations; 514 with complete ratings).
  Intervals are 95\% student-cluster bootstrap confidence intervals; $p_{\mathrm{Holm}}$ values are adjusted within research-question families.
  Conjunction tests have no single effect size; component estimates and intervals appear in \Cref{tab:conjunction-components}.}
  \label{tab:primary-results}
  \begin{tabular}{llrlrr}
    \toprule
    Primary claim & Outcome & Estimate & 95\% CI & $p$ & $p_{\mathrm{Holm}}$ \\
    \midrule
    $\rho(\mathcal{C}^\mathsf{C})$ %
    & \DIRQ & $-$0.079 & [$-$0.170, 0.011] & 0.959 & 1.000 \\
    & \TaSFC & $-$0.027 & [$-$0.121, 0.070] & 0.718 & 1.000 \\
    & \COH & $-$0.039 & [$-$0.128, 0.045] & 0.788 & 1.000 \\
    $\rho(\mathcal{C}^\mathsf{C}+\mathcal{W})-\rho(\mathcal{W})$ %
    & \DIRQ & $-$0.043 & [$-$0.102, 0.012] & 0.184 & 1.000 \\
    & \TaSFC & $-$0.049 & [$-$0.138, 0.040] & 0.332 & 1.000 \\
    & \COH & $-$0.019 & [$-$0.062, 0.026] & 0.431 & 1.000 \\
    Order conjunction & \DIRQ &  &  & 1.000 & 1.000 \\
    & \TaSFC &  &  & 0.268 & 0.803 \\
    & \COH &  &  & 1.000 & 1.000 \\
    Relation conjunction & \DIRQ &  &  & 1.000 & 1.000 \\
    & \TaSFC &  &  & 1.000 & 1.000 \\
    & \COH &  &  & 0.568 & 1.000 \\
    \bottomrule
  \end{tabular}
\end{table}

\begin{table}[t]
  \centering
    \caption{%
    Component contrasts for order and relational conjunction tests in $\mathsf{C}$ at Tier~3, using Coh-Metrix features and random forests.
    Conjunctions require both components to be positive; the joint $p$-value equals the larger component $p$-value if both are positive, and $1$ otherwise.
    \textit{Order vs.\ unordered} compares $\mathcal{C}^\mathsf{C}+\mathcal{V}^\mathsf{C}+\mathcal{O}^\mathsf{C}$ with $\mathcal{C}^\mathsf{C}+\mathcal{V}^\mathsf{C}$; \textit{order vs.\ permutations} uses $\mathcal{C}^\mathsf{C}+\mathcal{V}^\mathsf{C}+\widetilde{\mathcal{O}}^\mathsf{C}$ as control.
    \textit{Relations beyond content and order} compares $\mathcal{C}^\mathsf{C}+\mathcal{V}^\mathsf{C}+\mathcal{O}^\mathsf{C}+\mathcal{G}^\mathsf{C}$ with $\mathcal{C}^\mathsf{C}+\mathcal{V}^\mathsf{C}+\mathcal{O}^\mathsf{C}$; \textit{observed vs.\ randomized relations} uses $\mathcal{C}^\mathsf{C}+\mathcal{V}^\mathsf{C}+\mathcal{O}^\mathsf{C}+\widetilde{\mathcal{G}}^\mathsf{C}$ as control.
    Intervals are 95\% student-cluster bootstrap confidence intervals; component $p$-values are unadjusted.}
    \label{tab:conjunction-components}
  \small
  \begin{tabular}{lllrlr}
    \toprule
    RQ & Outcome & Component & $\Delta\rho$ & 95\% CI & $p$ \\
    \midrule
    RQ3 & \DIRQ & Order vs.\ unordered & $-0.032$ & [$-0.092$, $+0.028$] & 0.396 \\
     &  & Order vs.\ permutations & $-0.032$ & [$-0.099$, $+0.040$] & 0.426 \\
    RQ3 & \TaSFC & Order vs.\ unordered & $+0.098$ & [$+0.025$, $+0.166$] & 0.008 \\
     &  & Order vs.\ permutations & $+0.031$ & [$-0.024$, $+0.085$] & 0.268 \\
    RQ3 & \COH & Order vs.\ unordered & $-0.035$ & [$-0.112$, $+0.041$] & 0.413 \\
     &  & Order vs.\ permutations & $-0.037$ & [$-0.118$, $+0.046$] & 0.380 \\
    \midrule
    RQ4 & \DIRQ & Relations beyond content and order & $+0.017$ & [$-0.038$, $+0.075$] & 0.594 \\
     &  & Observed vs.\ randomized relations & $-0.031$ & [$-0.070$, $+0.003$] & 0.100 \\
    RQ4 & \TaSFC & Relations beyond content and order & $-0.033$ & [$-0.087$, $+0.020$] & 0.301 \\
     &  & Observed vs.\ randomized relations & $-0.026$ & [$-0.056$, $+0.002$] & 0.095 \\
    RQ4 & \COH & Relations beyond content and order & $+0.017$ & [$-0.036$, $+0.072$] & 0.568 \\
     &  & Observed vs.\ randomized relations & $+0.041$ & [$-0.003$, $+0.083$] & 0.073 \\
    \bottomrule
  \end{tabular}
\end{table}

\Rfinal{%
Non-significance does not establish absence of an effect.
Post hoc, one-sided 95\% upper bounds were at most $0.16$ for six RQ2 contrasts and $0.05$--$0.16$ for twelve conjunction components; a replication would have approximately 80\% power to detect a correlation of $0.13$ or an increment of $0.08$ in a single unadjusted test.}

\begin{figure}[ht]
    \centering
    \includegraphics[width=0.9\textwidth]{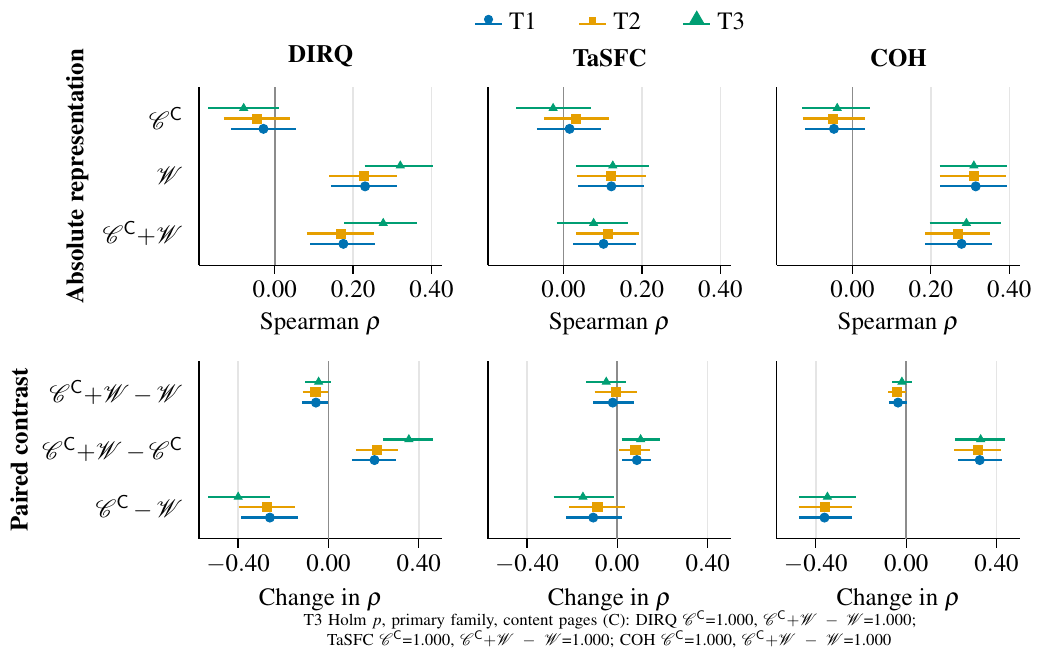}
    \caption{%
    RQ1--RQ2 results for content-page set $\mathsf{C}$.
    The upper panel shows within-task Spearman correlations for consulted-page content ($\mathcal{C}^\mathsf{C}$), written-response features ($\mathcal{W}$), and their concatenation ($\mathcal{C}^\mathsf{C}+\mathcal{W}$) across three validation tiers.
    The lower panel shows paired correlation differences, positive when the first representation performs better.
    Error bars indicate 95\% student-cluster bootstrap confidence intervals.
    Primary Tier~3 quantities are $\rho(\mathcal{C}^\mathsf{C})$ and $\rho(\mathcal{C}^\mathsf{C}+\mathcal{W})-\rho(\mathcal{W})$; other comparisons are exploratory.
    }
    \label{fig:rq1}
\end{figure}

For \Rfinal{RQ2}, the seventeen Coh-Metrix indices of consulted content pages showed no significant within-task predictive performance on held-out tasks.
Within-task correlations for $\mathcal{C}^\mathsf{C}$ were $-0.079$ for \DIRQ, $-0.027$ for \TaSFC, and $-0.039$ for \COH, with all 95\% confidence intervals including zero (\Cref{fig:rq1}).
Written-response features performed better ($\rho=0.320$, $0.126$, and $0.310$, respectively), with all correlations significant before correction.
Adding consulted-page features to written-response features yielded no improvement: $\rho(\mathcal{C}^\mathsf{C}+\mathcal{W})-\rho(\mathcal{W})$ was $-0.043$, $-0.049$, and $-0.019$, respectively.
Conversely, adding written-response features to consulted-page features %
improved prediction for \DIRQ ($0.356$) and \COH ($0.330$), both $p_{\mathrm{Holm}}=0.001$ in the exploratory family.
The improvement was smaller for \TaSFC ($0.103$, $p_{\mathrm{Holm}}=0.105$).
The larger contrast for outcomes assessing written responses suggests shared-source overlap (\Cref{sec:rq1}), not additional predictive information from consulted-page content.

For RQ3, no order conjunction test was significant (\Cref{tab:conjunction-components}).
Only \TaSFC showed a significant gain from adding observed order to the order-invariant baseline ($0.098$, $p=0.008$), but not over the permutation control ($0.031$, $p=0.268$).
The conjunction was therefore not significant ($p=0.268$).
Thus, the \TaSFC gain cannot be attributed specifically to observed consultation order.
For \DIRQ and \COH, both component estimates were negative.

For RQ4, no relational conjunction test was significant.
The largest positive component was for \COH against randomized relations ($\Delta\rho=0.041$, $p=0.073$), versus only $0.017$ ($p=0.568$) against the content-and-order baseline.
No other component contrast had a positive estimate exceeding $0.02$.

Secondary and exploratory outcomes showed similar results in the primary setting.
None of the 24 trace-representation correlations for Overall, the two \DIRQ facets, and task-standardized outcomes survived correction; the largest was $0.091$ for Overall with observed order ($p_{\mathrm{Holm}}=0.386$).
No conjunction was significant (minimum adjusted $p=0.902$).
However, sentence embeddings yielded significant correlations with task-standardized \DIRQ (\Cref{sec:source-results}): $\rho=0.138$ for the order-invariant baseline and $0.139$ for the full representation (both $p_{\mathrm{Holm}}=0.024$).
\Rfinal{Results in page sets $\mathsf{A}$ and $\mathsf{S}$ resembled those in $\mathsf{C}$, and excluding an unusually severe \COH rater changed within-task correlations by at most $0.02$.
Across both feature sources (792 exploratory conjunction tests each), only three conjunctions survived correction, all relational and for source referencing in $\mathsf{A}$ at Tiers~1--2; none occurred at Tier~3.}

\subsection{Incremental predictive information and the validation tiers}\label{sec:increment-results}

\begin{figure}[t]
    \centering
    \includegraphics[width=0.9\textwidth]{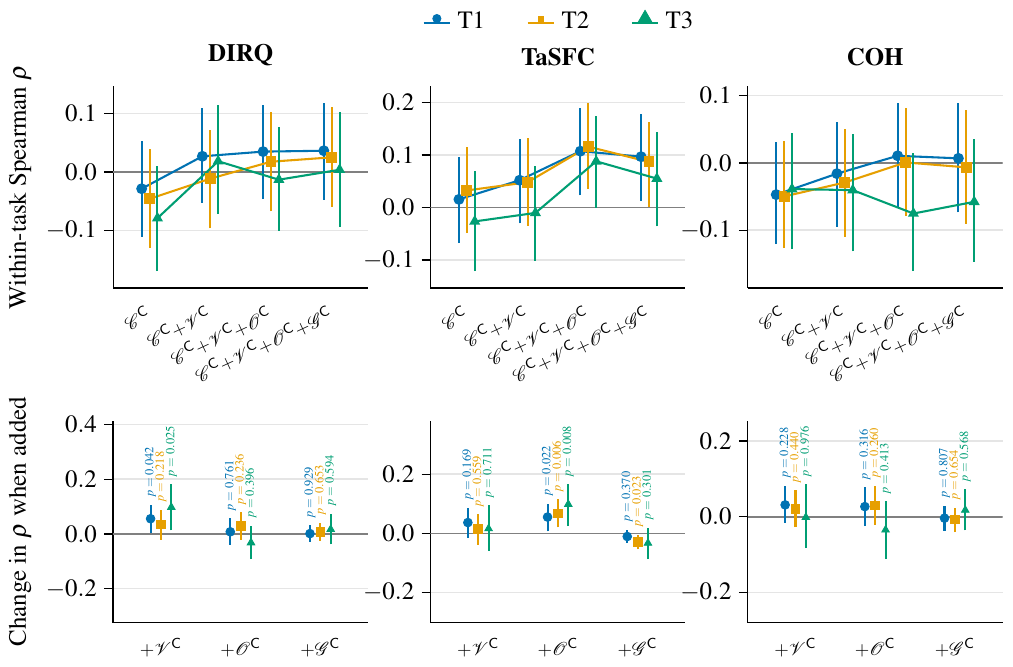}
    \caption{%
    Incremental predictive information for page set $\mathsf{C}$ across four cumulative specifications: content, visit descriptors, order-sensitive features, and relational features.
    The upper panel shows within-task Spearman correlations across three validation tiers.
    The lower panel shows successive correlation increments and unadjusted permutation $p$-values; Holm-adjusted values appear in the text.
    All specifications use identical student-task observations within each outcome.}
    \label{fig:cross-rq}
\end{figure}

Because all research questions used identical observations within $\mathsf{C}$, the four fixed-vector specifications form a progression of increasingly complex representations (\Cref{fig:cross-rq}).
At Tier~3, two increments, one for \DIRQ and one for \TaSFC, were statistically significant before %
correction; across all three tiers, six increments met this criterion.
Adding visit descriptors increased \DIRQ's correlation by $0.098$ (95\% CI $[0.013,0.181]$, $p=0.025$), narrowly missing Holm significance ($p_{\mathrm{Holm}}=0.051$).
Adding order-sensitive features increased \TaSFC's correlation by $0.098$ ($p=0.008$, exploratory $p_{\mathrm{Holm}}=0.017$).
\Rfinal{As reported in \Cref{sec:rq-results}, the corresponding conjunction was not significant.}
Relational increments remained within $\pm0.05$ across outcomes and tiers.

\Rfinal{For RQ5,} performance did not decline monotonically from Tier~1 to Tier~3.
None of the 36 tier comparisons for the four trace representations survived correction.
Five other comparisons were significant, all for \DIRQ: written-response and combined document-response features performed better at Tier~3 than at Tier~1 (differences $-0.090$, $p_{\mathrm{Holm}}=0.014$, and $-0.102$, $p_{\mathrm{Holm}}=0.008$) or Tier~2 ($-0.093$ and $-0.107$, both $p_{\mathrm{Holm}}=0.008$).
Permuted-order features also differed between Tiers~1 and~2 ($0.040$, $p_{\mathrm{Holm}}=0.048$).
None of twelve prespecified overstatement tests supported a Tier~1 or Tier~2 advantage over Tier~3 exceeding $0.05$ (maximum observed difference $0.062$, all $p\geq0.41$).
Thus, less restrictive splits showed no evidence of systematically inflated within-task correlations for trace representations.

\subsection{\Rfinal{Pooling gaps}}\label{sec:task-recognition-results}

\begin{figure}[t]
    \centering
    \includegraphics[width=0.85\textwidth]{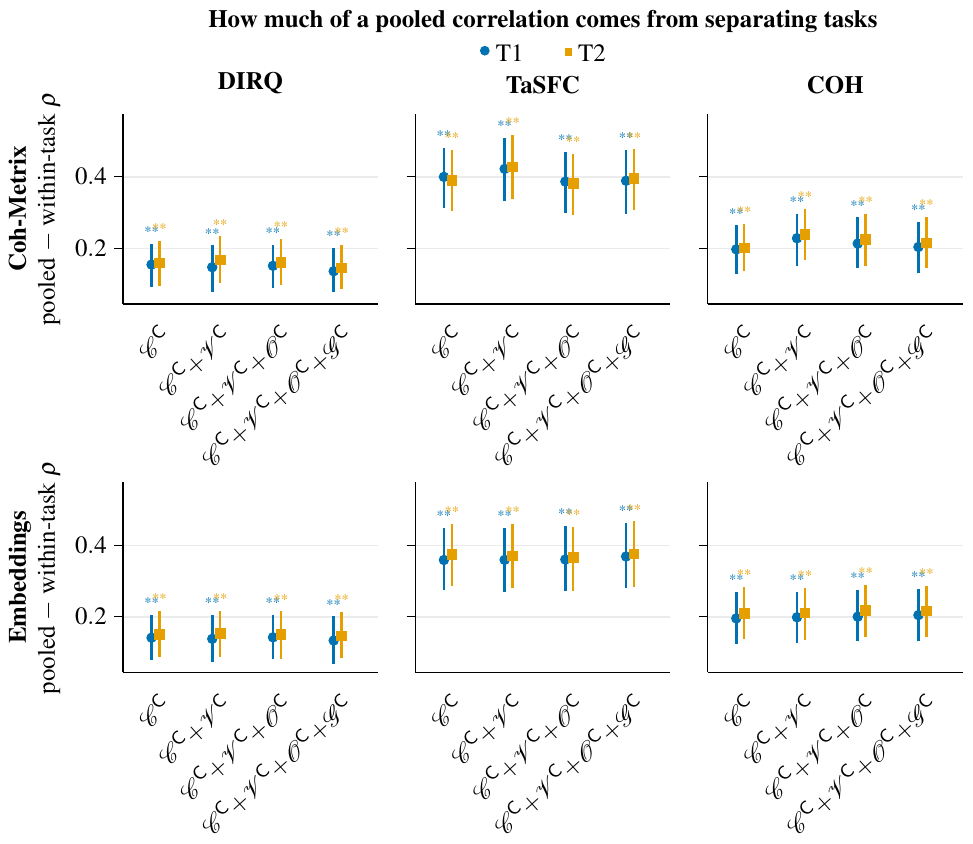}
    \caption{%
    Pooling gaps for page set $\mathsf{C}$ at Tiers~1--2, defined as pooled minus within-task Spearman correlations from identical out-of-fold predictions.
    Both feature sources, four cumulative specifications, and three primary outcomes are shown.
    Error bars show 95\% student-cluster bootstrap confidence intervals.
    Positive gaps indicate stronger pooled associations, consistent with contributions from task-level differences.}
    \label{fig:task-recognition}
\end{figure}

Previous correlations averaged within-task student rankings; pooling predictions and ratings across tasks instead reveals associations attributable to task differences.
The pooling gap measures this difference without changing predictions, repeats, or seeds \Rfinal{(\Cref{fig:task-recognition})}.
This prespecified exploratory comparison covers Tiers~1 and~2 and lies outside the confirmatory families.
For page content ($\mathcal{C}^\mathsf{C}$) with random forests at Tier~1, the pooling gaps were $0.400$ for \TaSFC, $0.197$ for \COH, and $0.154$ for \DIRQ.
Comparable gaps occurred at Tier~2 and across fixed-vector specifications.
Across both tiers, estimators, specifications, and unstandardized outcomes in $\mathsf{C}$, 525 of 576 Coh-Metrix pooling gaps were significantly positive after correction (median $0.169$).
The pooling gaps were smaller for written-response features: $0.132$ for \TaSFC, $0.059$ for \DIRQ, and $0.056$ for \COH.
These features describe the rated product rather than browsing traces.
At Tier~1, consulted-page content yielded pooled versus within-task correlations of $0.415$ versus $0.015$ for \TaSFC, $0.126$ versus $-0.029$ for \DIRQ, and $0.150$ versus $-0.047$ for \COH.
These gaps suggest that task differences contributed more to pooled associations than within-task student ranking.

\Rfinal{As expected once task-level outcome differences are removed, training and evaluating on within-task standardized outcomes reduced the median pooling gap from $0.169$ to $0.008$.}
None of the 144 Coh-Metrix pooling gaps remained significantly positive after correction.
For page content ($\mathcal{C}^\mathsf{C}$) at Tier~1, the gap fell from $0.400$ to $0.024$ for \TaSFC and from $0.154$ to $0.001$ for \DIRQ.
By contrast, the Tier~3 within-task correlation for written-response predictions of \DIRQ remained nearly unchanged ($0.320$ vs.\ $0.327$; \Cref{tab:task-standardized-results}).
Standardization removed task-level differences in outcome means and variances while preserving within-task rankings.

\subsection{The second feature source and the neural models}\label{sec:source-results}

\begin{figure}[t]
    \centering
    \includegraphics[width=0.8\textwidth]{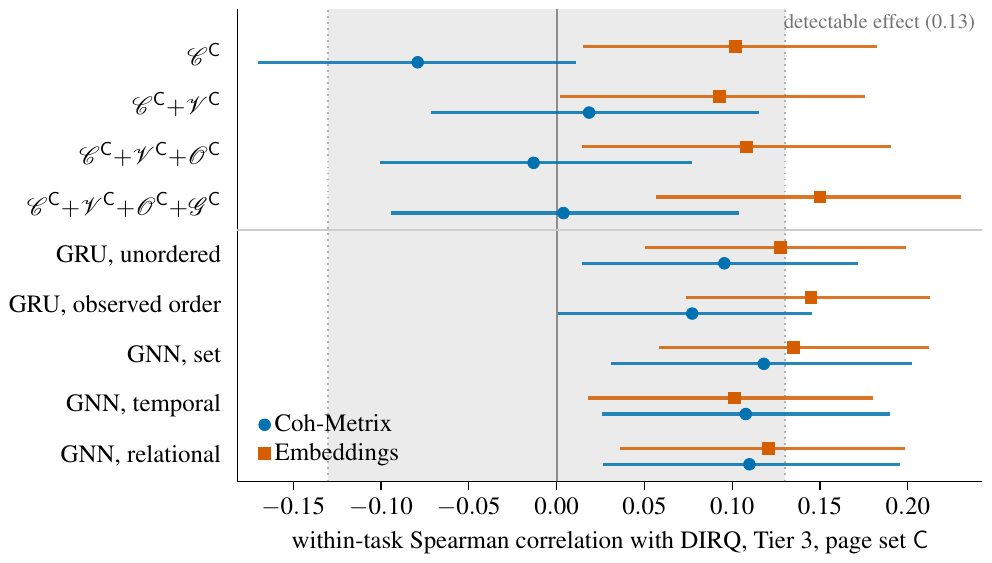}
    \caption{%
    Within-task Spearman correlations with \DIRQ at Tier~3 for page set $\mathsf{C}$, by representation and feature source, with 95\% student-cluster bootstrap confidence intervals.
    The shaded band marks effects below the approximate 80\%-power detection threshold ($|\rho|<0.13$).
    The horizontal line separates random-forest fixed-vector specifications from neural models.
    Matched contrasts show no significant incremental contribution from observed order or relational structure.}
    \label{fig:source-comparison}
\end{figure}

For \DIRQ, sentence embeddings generally outperformed Coh-Metrix features (\Cref{fig:source-comparison}).
At Tier~3 in $\mathsf{C}$, embedding-based trace correlations ranged from $\rho=0.09$ to $0.15$, versus $-0.079$ to $0.019$ for Coh-Metrix.
The full specification, $\mathcal{C}^\mathsf{C}+\mathcal{V}^\mathsf{C}+\mathcal{O}^\mathsf{C}+\mathcal{G}^\mathsf{C}$, reached $\rho=0.150$ (95\% CI $[0.056,0.230]$, $p_{\mathrm{Holm}}=0.024$).
Consulted-page content ($\mathcal{C}^\mathsf{C}$) alone reached $\rho=0.102$ (95\% CI $[0.015,0.182]$, $p=0.014$, $p_{\mathrm{Holm}}=0.115$).
Combining consulted-page content with written-response features yielded $\rho=0.182$ ($p_{\mathrm{Holm}}=0.006$), versus $0.174$ for responses alone. The $0.008$ increment provides little evidence of additional predictive information from consulted pages.
The embedding advantage was confined to \DIRQ: the full specification yielded $-0.014$ for \TaSFC and $0.008$ for \COH, with similarly small Coh-Metrix estimates.
Embedding correlations of $0.09$--$0.15$ bracket the approximately $0.13$ detectable with 80\% power in the primary analysis.
Thus, both feature sources are compatible with a modest association captured more effectively by embeddings.

\Rfinal{%
Across \numprint{6192} within-task correlations, all values above $0.4$ involved written responses; trace-only correlations peaked at $0.246$ ($0.223$ at Tier~3), with similar distributions for observed structures and matched controls.
With embeddings, \TaSFC correlations of $0.16$--$0.31$ at Tiers~1--2 dropped to $-0.01$--$0.10$ at Tier~3 (\Cref{tab:response-spearman}); the two highest trace-only correlations ($0.246$ and $0.227$ for task-standardized \TaSFC) fell to $-0.019$ and $0.003$, respectively.
}

\section{Discussion}\label{sec:Discussion}
We investigated whether consulted-page content, visit order, and relational organization predict ratings of students' written responses.
This required within-task prediction, transfer to unseen tasks, and, for structural features, gains over matched permutation controls.
Each analysis yielded a distinct finding.
Across 576 configurations at Tiers~1 and~2, pooled correlations exceeded within-task correlations by a median of $0.169$.
The criterion audit qualified null findings: \TaSFC's stable between-student variance was only 8\% and not distinguishable from zero, making null results less informative than for \DIRQ or \COH.
Finally, permutation controls tested whether predictive gains reflected observed order or relational structure rather than additional features.

The prespecified results provide little support for incremental predictive contributions.
At Tier~3, Coh-Metrix features of consulted-page content showed no significant within-task association with performance, and neither structural conjunction test was significant.
For \TaSFC, observed-order features improved prediction over the order-invariant baseline ($\Delta\rho=0.098$, exploratory $p_{\mathrm{Holm}}=0.017$), but not significantly over permuted order ($\Delta\rho=0.031$, $p=0.268$), leaving the conjunction non-significant.
The prespecified embedding sensitivity analysis yielded small, significant positive correlations with \DIRQ, but matched comparisons did not establish contributions from observed order or relational structure.
Thus, absolute predictive associations did not establish incremental value from specific aspects of consultation behavior.

\subsection{What a pooled correlation measures}
At Tier~3, the primary Coh-Metrix--random-forest specification yielded within-task correlations of $-0.079$ for \DIRQ, $-0.027$ for \TaSFC, and $-0.039$ for \COH, all with confidence intervals including zero.
This does not rule out predictive associations: post-hoc sensitivity analysis suggests correlations below $0.13$ would often go undetected in individual unadjusted tests, with still lower power after Holm correction.

Pooling predictions across tasks yielded substantially higher correlations than within-task evaluation.
Across 576 Coh-Metrix configurations in $\mathsf{C}$, 525 pooling gaps were significantly positive after correction (median $0.169$).
Because configurations share observations, this indicates robustness across analytical choices, not independent replication.
For Tier~1 consulted-page content predicting \TaSFC, pooled and within-task correlations were $0.415$ and $0.015$, respectively.
\Rfinal{As expected, task-wise outcome standardization removed the gap (median $0.008$).}
Thus, task differences contributed substantially to pooled associations, consistent with models capturing task-related variation in traces and ratings without necessarily identifying tasks explicitly.

The validation tiers address a different question \Rfinal{(RQ5)}.
None of the twelve prespecified overstatement tests established that Tier~1 or Tier~2 outperformed Tier~3 by more than $0.05$.
None of the 36 Holm-adjusted tier comparisons for the four observed trace specifications was significant.
\Rfinal{Written-response features even predicted \DIRQ better at Tier~3 than at Tiers~1 and~2.}
Thus, withholding an entire task does not necessarily reduce within-task correlations.
Validation tiers differ in what is withheld, not in a presumed monotonic decline in performance.
Why written-response features predicted \DIRQ better at Tier~3 remains unclear.

\subsection{\Rfinal{What the criterion audit implies}}

Non-significant predictions may reflect \Rfinal{weak predictors, an unstable criterion,} or both, particularly for \TaSFC.
Student-by-task variation accounted for 64\% of \TaSFC variance (95\% profile-likelihood interval: 53\%--69\%), without distinguishing task from occasion effects.
Stable between-student variance was 8\% (95\% interval: 0\%--20\%), and relative generalizability for one task rated by three raters was $0.10$.

\TaSFC may still be predictable on a held-out task, but the audit limits its interpretation as a stable, task-general attribute.
Its four non-significant tests are therefore less informative about task-general predictive validity than those for \DIRQ and \COH.
Stable between-student variance was 23\% for \DIRQ and 19\% for \COH, with single-task generalizability coefficients of $0.39$ and $0.40$, respectively.
Both components differed significantly from zero, but their magnitudes cannot be compared precisely given the confidence intervals and few repeated students.
For \DIRQ, student-by-task variance was also 23\%, leaving the relative contributions uncertain.
Thus, non-significant \DIRQ and \COH results more strongly, though still cautiously, constrain the task-general predictive value of the tested trace representations.

Rater effects warrant similar caution.
Estimated rater variance was 2\% for \TaSFC and 19\% for \COH, but both estimates are imprecise given six raters with unequal workloads.
Systematic rater severity differences were established for \COH, without evidence that the effects differ significantly between outcomes.

\subsection{Where positive associations appeared}
Sentence embeddings yielded the strongest trace-based associations with performance.
At Tier~3, the full fixed-vector specification correlated with \DIRQ at $\rho=0.150$, significant after correction; seven of eight neural specifications were also significant.
However, five were unordered baselines or randomized controls, some outperforming their observed counterparts.
No matched fixed-vector, graph, or sequence comparison established significant gains from observed order or relational structure.
Thus, embeddings predicted \DIRQ without identifying which trace aspects contributed.
Comparable evidence was absent for \TaSFC and \COH.

Written-response features showed the strongest within-task association with \DIRQ ($\rho=0.320$).
Adding \Rfinal{Coh-Metrix page} features did not improve prediction beyond the written response alone.
Because features and ratings derived from the same written text, this association does not establish predictive value for the preceding browsing process.

The rated response contained performance-related information beyond the seventeen Coh-Metrix indices.
Within tasks, response length correlated with \DIRQ ($\rho=0.391$) and \COH ($\rho=0.338$), remaining nearly unchanged after controlling for these indices ($0.393$ and $0.337$).
Length was not among the indices.
These associations do not establish that length contributed to the response model's predictive performance.

\subsection{Implications for trace-based assessment}
The findings support five recommendations for evaluating trace-based assessments.
First, report pooled and within-task performance from identical out-of-fold predictions when multiple tasks are involved.
Their difference indicates the contribution of between-task variation without constituting an additive decomposition.
This extends concerns about aggregated metrics in learning analytics \cite{Gardner:et:al:2018,Arizmendi:et:al:2023} by highlighting task boundaries as an evaluation choice.

Second, claims of transfer to unseen tasks require evaluation on tasks excluded from training.
Our study tests transfer across tasks within one multisite assessment, whereas earlier work examined transfer across institutions using a shared instrument \cite{Xu:et:al:2026}.
These designs address distinct generalization boundaries.
Positive embedding-based associations with \DIRQ indicate cross-task predictive transfer, but matched comparisons established no additional contribution from observed order or relational structure.

Third, claims about order and relational structure require controls that reorganize the same events while preserving content and model class.
Sequence-performance associations alone do not establish order-specific predictive information \cite{Yu:et:al:2025}.
In the present study, several observed-order models correlated positively with \DIRQ, but shuffled or randomized controls performed similarly, with no significant structural advantage after correction.

Fourth, assessment tasks and session contents must be documented to interpret browsing traces as evidence about students.
Task prompts constrain consulted pages, written responses, and evaluation criteria.
Search-results pages, content pages, and responses capture distinct aspects of this process.
Trace interpretations must therefore account for both the task and the underlying page set.

Finally, model evaluation cannot compensate for a criterion lacking task-general validity.
Task-general interpretation requires stable between-student differences, which the criterion audit did not establish for \TaSFC.
Single-task prediction therefore cannot validate \TaSFC as a task-general attribute, and averaging across tasks improves generalizability only if such stability exists.

\subsection{Limitations and future directions}
This retrospective study evaluates one multisite assessment without prospective or external validation.
The final plan was fixed before modeling corrected data but informed by earlier analyses, precluding independent preregistration.
Tier~3 holds out each of four tasks in turn, the smallest containing 86 scored observations.
Moreover, the inferential intervals resample students rather than tasks and therefore do not quantify uncertainty arising from the selection of these four tasks.
Holding out different tasks leaves 337--441 scored training observations, confounding task identity with training-set size.
Thus, transfer findings apply to the studied tasks, not an unrestricted population of future tasks.

The primary analysis includes only student-task observations with at least two analyzable content pages.
Students consulting fewer than two pages may differ in information seeking or performance but lacked eligible browsing traces.
\Rfinal{Ratings, however, did not differ systematically between responses with and without at least two analyzable pages (mean \DIRQ 0.609 vs.\ 0.594, \TaSFC 0.457 vs.\ 0.469, \COH 0.655 vs.\ 0.649). None of the twelve task-wise comparisons approached significance (all $p\geq0.135$).}
Of eligible traces, 10\% contained exactly two pages, for which RQ3 captures only visit direction and the RQ4 observed-versus-randomized contrast is structurally zero.
Thus, RQ3 and RQ4 estimate average increments including traces with limited or no structural variation.

URL-based page classification makes content pages a proxy for documents.
Coh-Metrix indices use a German adaptation; topic, genre, and emotion classifiers were not validated on these pages.
The representations omit queries, dwell times, within-page reading, source credibility, and task-specific relevance.
Thus, non-significant findings concern the tested representations and effect sizes, not browsing traces generally.

The criterion audit includes only 136 students assessed once on each of two tasks, limiting evidence of stable between-student variance, particularly for \TaSFC.

The study covers four tasks, one assessment framework and construct, three universities, and one national context.
This reflects the context dependence reported by \citeA{Gasevic:et:al:2016}, who found varying trace-feature contributions across nine courses.
Predictive value depends on recorded behavior, performance measures, and training contexts.
Replication should therefore vary tasks, frameworks, institutions, and populations, not merely algorithms.

A key objective for future research is to develop representations that generalize across tasks while remaining sensitive to meaningful differences between students.
Future studies could compare selected with available sources or model alignment between consulted content and task demands, capturing information seeking on a common, task-relative scale.
However, features must retain consistent meanings and comparable predictive relationships across tasks.
This requires diverse tasks, repeated student observations, and explicit tests of cross-task comparability and transfer.

\subsection{Conclusion}
Our study shows that high correlations between browsing traces and performance may largely reflect differences between tasks rather than between students within tasks.
Across 576 configurations, pooled correlations exceeded within-task correlations by a median of $0.169$.
This suggests models captured task-related differences without necessarily recognizing task identity.

On held-out tasks, \Rfinal{Coh-Metrix page features} showed no reliable within-task prediction, nor did observed order or relational structure add significant predictive value.
Observed-order features improved \TaSFC prediction over the unordered baseline but not over permuted order, precluding attribution to the actual consultation sequence.
Sentence embeddings yielded modest correlations with \DIRQ (approximately $0.15$), unlike Coh-Metrix features.
However, unordered and randomized controls showed similar associations, with no significant structural advantage.
Thus, traces predicted \DIRQ modestly, but the responsible features remain unidentified.

In the primary analysis, written-response features showed the strongest within-task association with \DIRQ ($\rho=0.32$).
Because predictors and ratings derive from the same text, this does not independently implicate browsing behavior.

Non-significant results do not rule out predictive value in browsing traces or their structure.
Power for small associations was limited, stable between-student variance was not established for \TaSFC, short traces constrained structural comparisons, and transfer involved only four tasks.

Three evaluation checks distinguished apparent performance from generalizable predictive information.
Comparing pooled and within-task correlations revealed the contribution of task differences.
\Rfinal{Criterion audits checked whether stable between-student variation supported task-general interpretations.}
Matched controls tested whether observed order or relational structure added predictive value beyond reorganized traces.
These findings have implications beyond the present assessment:
predicting ratings, distinguishing students within tasks, generalizing to unseen tasks, and identifying the predictive contribution of specific trace properties are distinct achievements that require distinct evidence.
\phantomsection
\section*{Ethical Considerations}
\addcontentsline{toc}{section}{Ethical Considerations}
\Rfinal{
This study uses students' browsing histories and written responses to model performance on online critical reasoning tasks.
Because these data may reveal sensitive information about students' behavior and achievement, their collection and use require informed consent, privacy safeguards, and secure data handling.
We also note that models trained on such data may reproduce existing biases and may not generalize across prompts, student populations, or educational settings.
Accordingly, we do not consider these models appropriate for high-stakes decisions about individual students, and any educational use should remain low-stakes and subject to human oversight.
Finally, because browsing behavior is only an indirect proxy for reading and reasoning, model outputs should not be interpreted as direct evidence of students' comprehension strategies or cognitive states.
}
\phantomsection

\section*{Declaration of Conflicting Interest} %

\addcontentsline{toc}{section}{Declaration of Conflicting Interest} %

The author(s) declared no potential conflicts of interest with respect to the research, authorship, and/or publication of this article.

\phantomsection
\printbibliography

\clearpage
\begin{refsection}
\setcounter{table}{0}\setcounter{figure}{0}
\renewcommand{\thetable}{S\arabic{table}}\def\theHtable{S\arabic{table}}
\renewcommand{\thefigure}{S\arabic{figure}}\def\theHfigure{S\arabic{figure}}
\phantomsection\addcontentsline{toc}{section}{Supplementary Material}
\begin{center}
{\bfseries\Large Supplementary Material}\par
\end{center}
\bigskip
\subsection*{Data reconstruction and trace audit}

\Rfinal{During task completion, page openings were logged and the text of each opened page was captured.
The capture could store several versions of the text associated with one visit.}
Each visit received the largest usable version, and the latest version when several were equally large, because late captures were sometimes partial loads or consent overlays.
Pages captured without a separate opening event were inserted at their first capture time.
Assessment-platform pages, browser-internal and non-web addresses, and search-engine click-through redirects were removed.
Successive openings of the same URL within two seconds were treated as one load, whereas later returns were retained as revisits; dwell time was not used.

\Rfinal{The 50-word minimum for analyzable texts (\Cref{sec:population})} is not required by Coh-Metrix; in our data, shorter captures often consisted of navigation fragments or isolated sentences, for which adjacent-sentence indices are undefined \cite{Graesser:et:al:2004}.

Named-source URLs were matched to visits at page or site level after resolving search-engine redirects, removing protocols, leading \code{www.} or \code{m.}, query strings, and trailing slashes, and lower-casing the host and path.

Topic, genre, and emotion labels were assigned with \textit{classla/multilingual-IPTC-news-topic-classifier},
\textit{classla/xlm-roberta-base-multilingual-text-genre-classifier}, and \textit{MilaNLProc/xlm-emo-t}, respectively.

\paragraph{Audit procedures.}\label{par:Audit procedures}
The audit tracked student-task observations at successive stages of data processing, from rater scores and process observables to eligibility and feature extraction, while examining page coverage and named-source recovery.
A named source was classified as \textit{opened} if its corresponding page appeared in the same student-task session, and as analyzed if its textual content was included in the analysis.
The audit used outcome data solely to determine the availability of rater scores for each observation.

Among pages with recorded visits and captured text, 94\% were in German, compared with 89\% of content pages with captured text.

\begin{figure}[ht]
    \centering
    \includegraphics[width=\textwidth]{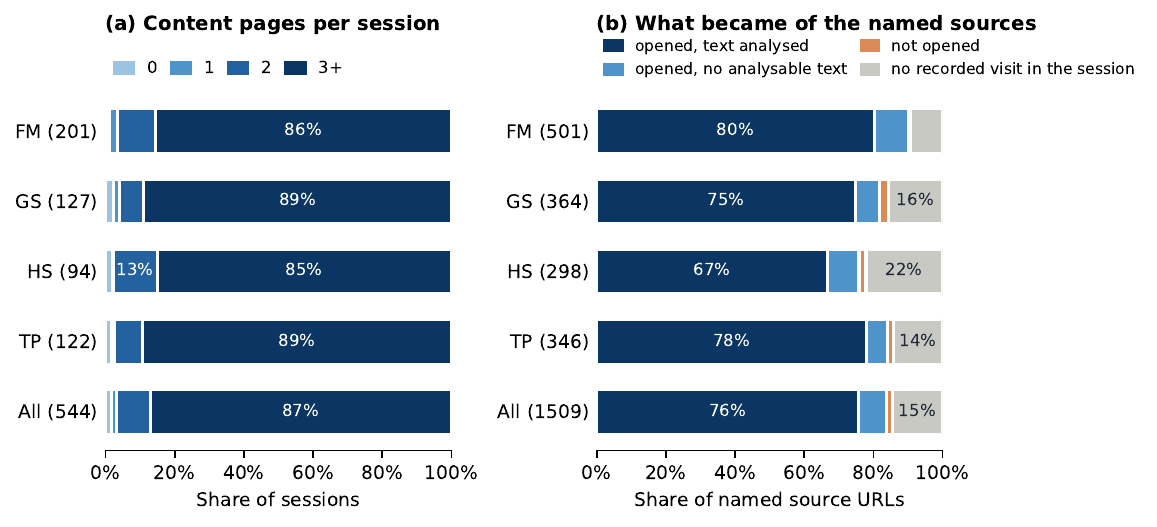}
    \caption{Trace audit. \Rfinal{FM, GS, HS, and TP denote the four tasks: health risks from cell-phone towers, poisoning from a traditional German dish, mosquito-bite treatment with heat sticks, and Tetra Pak recyclability; All combines them.} (a) Content pages with analyzable text per session and task; parentheses give the sessions with a recorded visit. (b) Source URLs named in the answers, classified as opened and analyzed, opened without analyzable text, not opened, or belonging to a session without a recorded visit\Rfinal{; parentheses give the number of named URLs}.}
    \label{fig:trace-audit}
\end{figure}

Content pages were longer than SERPs (median: 759 vs. 220 words; \Rfinal{\Cref{tab:text-lengths}}).
Most sessions with recorded visits had at least three content pages with analyzable text (474 of 544 sessions; median 5).
We also assessed whether sources cited in students' responses matched recorded page visits.
Of these sources, 76\% matched recorded visits with analyzable text and 8\% matched visits without it.
Fewer than 2\% lacked matching visits despite available traces, while 15\% came from sessions without recorded visits (percentages rounded; \Rfinal{\Cref{fig:trace-audit}}).

\subsection*{Representation and model details}

\begin{table}[ht]
\centering
\caption{The analysis grid. Specifications are written without the \Rfinal{page-set superscript} because the same eight fixed specifications were fitted in each \Rfinal{page set}. Neural specifications were fitted only in \Rfinal{page set $\mathsf{C}$} and only for the three confirmatory outcomes. Task-standardized and weighted outcomes were fitted with the fixed estimators only, and the weighted outcomes only at Tier~3. Task recognition is defined at Tiers~1 and~2, so its cell counts are twice the per-tier totals: $11\cdot3\cdot8+3\cdot8=288$ unstandardized cells and $3\cdot3\cdot8=72$ task-standardized cells, giving the 576 and 144 gaps reported in \Rfinal{\Cref{sec:task-recognition-results}}.}
\label{tab:analysis-grid}
\small
\begin{tabular}{@{}llr@{}}
\toprule
Factor & Elements & $n$ \\
\midrule
Tier & T1, T2, T3 & 3 \\
Feature source & Coh-Metrix, embeddings & 2 \\
\Rfinal{Page set} & \Rfinal{$\mathsf{A}$, $\mathsf{C}$, $\mathsf{S}$} & 3 \\
Fixed estimator & Random Forest, SVR-RBF, PLS & 3 \\
\addlinespace
\multicolumn{3}{@{}l}{\emph{Specifications}} \\
Fixed & \Rfinal{$\mathcal{C}$;\; $\mathcal{W}$;\; $\mathcal{C}+\mathcal{W}$;\; $\mathcal{C}+\mathcal{V}$;} & 8 \\
 & \Rfinal{$\mathcal{C}+\mathcal{V}+\mathcal{O}$;\; $\mathcal{C}+\mathcal{V}+\widetilde{\mathcal{O}}$;} & \\
 & \Rfinal{$\mathcal{C}+\mathcal{V}+\mathcal{O}+\mathcal{G}$;\; $\mathcal{C}+\mathcal{V}+\mathcal{O}+\widetilde{\mathcal{G}}$} & \\
Neural & GRU unordered, observed, shuffled; & 8 \\
 & GNN set, temporal, temporal-shuffled, & \\
 & relational, randomized-relations & \\
\addlinespace
\multicolumn{3}{@{}l}{\emph{Outcomes}} \\
Confirmatory & \DIRQ, \TaSFC, \COH & 3 \\
Secondary & Overall & 1 \\
Exploratory facets & response fulfillment, referenced sources & 2 \\
Rater-rule rebuilds & \DIRQ and \TaSFC, each rebuilt with all raters and & 5 \\
 & without the severe rater; \COH without & \\
 & the severe rater & \\
\cmidrule(l){3-3}
\multicolumn{2}{@{}l}{\quad\emph{unstandardized outcomes combined}} & 11 \\
\addlinespace[1pt]
Task-standardized & \DIRQ, \TaSFC, Overall, each as a within-task $z$-score & 3 \\
Weighted & \DIRQ, \TaSFC & 2 \\
\bottomrule
\end{tabular}
\end{table}

\begin{table}[htbp]
\centering
\caption{Session descriptions (\Rfinal{$\mathcal{V}$}), order features (\Rfinal{$\mathcal{O}$}), and relation features (\Rfinal{$\mathcal{G}$}), as recorded by the representation adapter. Counts in parentheses; label shares are listed by the levels observed in the analyzed data, and a level absent from a training fold contributes no column there.}
\label{tab:feature-definitions}
\small
\begin{tabular}{llp{0.62\textwidth}}
\toprule
Block & Group & Features \\
\midrule
    \Rfinal{$\mathcal{V}$} & Session size and revisits (3) & log visits, log unique urls, revisit rate \\
    \Rfinal{$\mathcal{V}$} & Topic shares (16) & arts, culture, entertainment and media, conflict, war and peace, crime, law and justice, disaster, accident and emergency incident, economy, business and finance, education, environment, health, human interest, labour, lifestyle and leisure, politics, science and technology, society, sport, topic missing share \\
    \Rfinal{$\mathcal{V}$} & Genre shares (10) & Forum, Information/Explanation, Instruction, Legal, News, Opinion/Argumentation, Other, Promotion, Prose/Lyrical, genre missing share \\
    \Rfinal{$\mathcal{V}$} & Emotion shares (5) & anger, fear, joy, sadness, emotion missing share \\
    \Rfinal{$\mathcal{O}$} & Order geometry (16) & first pc1, first pc2, first pc3, first pc4, last pc1, last pc2, last pc3, last pc4, step rms mean, step rms std, step rms min, step rms max, endpoint rms, path efficiency, turn cosine mean, centroid distance change \\
    \Rfinal{$\mathcal{G}$} & Relation summaries (22) & topic pair same density, topic temporal same rate, topic transition entropy, genre pair same density, genre temporal same rate, genre transition entropy, emotion pair same density, emotion temporal same rate, emotion transition entropy, similarity mean, similarity std, similarity min, similarity max, weighted degree mean, weighted degree std, weighted degree max, temporal similarity mean, non temporal similarity mean, temporal similarity gap, temporal above median, laplacian gap, laplacian radius \\
\bottomrule
\end{tabular}
\end{table}

Short sequences constrain what the \Rfinal{order and relational} representations can express.
For a two-page sequence, the trajectory consists of a single step, so order information is limited to its direction.
The sole page pair is necessarily adjacent, leaving no distinction between adjacent and non-adjacent pairs.
Consequently, if the relational features are invariant under swapping the two pages, $\mathbf{g}_i^{\mathsf{U}}=\tilde{\mathbf{g}}_i^{\mathsf{U}}$.
The RQ3 and RQ4 estimates therefore include both trajectories with richer structural information and those with limited or absent structural contrasts.
\Cref{sec:results-sample} reports the prevalence of two-page sequences.

\begin{figure}[ht]
    \centering
    \includegraphics[width=\textwidth]{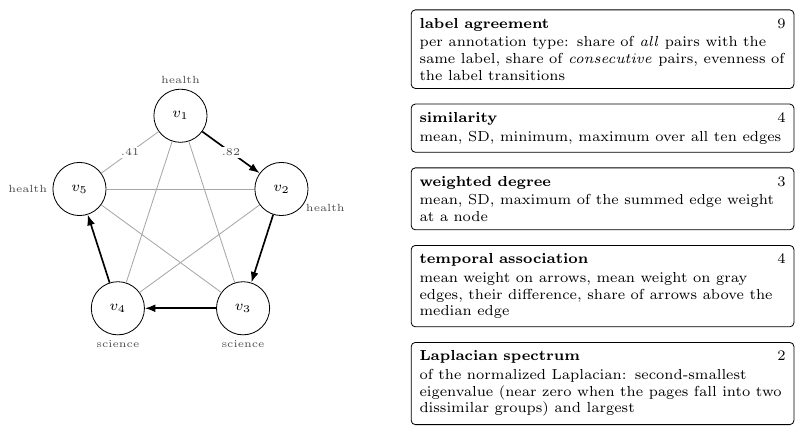}
    \caption{The relation block \Rfinal{$\mathcal{G}$}, illustrated for five visits. Every page pair is an edge weighted by the cosine similarity of its page vectors, rescaled to $[0,1]$; arrows identify temporally consecutive pairs. Labels and weights are illustrative. The control \Rfinal{$\widetilde{\mathcal{G}}$} recomputes the 22 summaries after jointly permuting visits, labels, and the similarity matrix. This preserves the weights and label composition but changes which relations align with temporal adjacency.}
    \label{fig:relation-features}
\end{figure}

Random Forest regression used 300 trees, a minimum leaf size of 5, and the square root of the feature count at each split.
Radial-basis support-vector regression used $C=1$, $\epsilon=0.1$, and the library-default $\gamma$, with the target standardized within the training fold.
Partial least-squares regression used two components.
The graph backbone comprised a node encoder, two rounds of edge-conditioned message passing with gated updates, and concatenated mean and maximum readouts followed by \Rfinal{$\mathcal{V}$}.
The GRU used one layer and its final state was concatenated with \Rfinal{$\mathcal{V}$}.
Both neural models used hidden dimension 32, learning rate .01, at most 50 epochs, and early stopping with patience 10 on a student-disjoint validation subset.
An estimator was labeled degenerate for a tier--source--outcome combination when more than 10\% of its block-level Spearman correlations were undefined.

Neural fits used three named seeds and ran in parallel on CPU.
A one-fold benchmark gave identical CPU and GPU predictions, and parallel-versus-serial numerical differences were negligible relative to between-seed variability.
Software versions, device, fit time, parameter count, and seed variability were retained in the run manifests.

In neural analyses, individual visits are represented as sequence elements or graph nodes containing the scaled page feature vector and, where applicable, normalized visit position.
Temporal edges connect successive visits, whereas relational graphs connect all ordered pairs of visits, with temporal adjacency, shared annotations, and cosine similarity encoded as edge attributes.
The neural readout produces a hidden-state vector $\mathbf{h}_i^\mathsf{U}$, which is concatenated with the order-invariant visit descriptors $\mathbf{v}_i^\mathsf{U}$ before prediction, ensuring that aggregate visit characteristics remain available independently of the learned sequential or relational representation.

For RQ4, we perform matched graph ablations to assess the incremental predictive contribution of relational information while controlling for differences in neural architecture.
These ablations use a shared message-passing backbone \cite{Gilmer:et:al:2017}, while the sequence benchmark for RQ3 uses a single-layer gated recurrent unit (GRU) \cite{Cho:et:al:2014}.\footnote{Permutation controls (see \Cref{sec:representations}) disrupt specific structural information, whereas matched graph ablations vary the available information while holding the model architecture fixed.}

Temporal graph models are compared with set-based graph models and graph models with shuffled visit order.
An observed-order GRU is compared with a GRU receiving a single fixed permutation of each visit sequence and an order-averaged GRU trained using eight fixed random permutations per sequence and evaluated by averaging predictions over 32 permutations.
The order-averaged GRU reduces sensitivity to the particular visit order but is not strictly permutation-invariant.

\begin{table}[t]
  \centering
    \caption{%
    Within-task Spearman correlations with \DIRQ at Tier~3 for page set $\mathsf{C}$, by feature source and representation.
    Fixed-vector specifications use random forests.
    The reported $p_{\mathrm{Holm}}$ values are adjusted within their respective test families.
    The content-only correlation $\rho(\mathcal{C}^\mathsf{C})$ changes sign between feature sources, but neither source provides statistically significant evidence of incremental predictive information attributable specifically to observed consultation order or relational structure.
  }
\label{tab:source-comparison}
  \small
  \begin{tabular}{lrrrr}
    \toprule
    & \multicolumn{2}{c}{Coh-Metrix} & \multicolumn{2}{c}{Embeddings} \\
    \cmidrule(lr){2-3}\cmidrule(lr){4-5}
    Representation & $\rho$ & $p_{\mathrm{Holm}}$ & $\rho$ & $p_{\mathrm{Holm}}$ \\
    \midrule
    $\mathcal{C}^\mathsf{C}$ %
    & $-0.079$ & $1.000$ & $+0.102$ & $0.115$ \\
    $\mathcal{C}^\mathsf{C}+\mathcal{V}^\mathsf{C}$
    & $+0.019$ & $1.000$ & $+0.093$ & $0.150$ \\
    $\mathcal{C}^\mathsf{C}+\mathcal{V}^\mathsf{C}+\mathcal{O}^\mathsf{C}$
    & $-0.013$ & $1.000$ & $+0.108$ & $0.088$ \\
    $\mathcal{C}^\mathsf{C}+\mathcal{V}^\mathsf{C}+\mathcal{O}^\mathsf{C}+\mathcal{G}^\mathsf{C}$
    & $+0.004$ & $1.000$ & $+0.150$ & $0.024$ \\
    GRU, unordered & $+0.095$ & $0.155$ & $+0.128$ & $0.006$ \\
    GRU, observed order & $+0.077$ & $0.480$ & $+0.145$ & $0.006$ \\
    GNN, set & $+0.118$ & $0.090$ & $+0.135$ & $0.010$ \\
    GNN, temporal & $+0.108$ & $0.114$ & $+0.101$ & $0.082$ \\
    GNN, relational & $+0.110$ & $0.110$ & $+0.121$ & $0.029$ \\
    \bottomrule
  \end{tabular}
\end{table}

The neural models also yielded positive predictive correlations with \DIRQ using both feature sources.
At Tier~3 in page set $\mathsf{C}$, all graph and sequence models produced positive within-task correlations with \DIRQ.
With Coh-Metrix features, estimates ranged from $0.058$ to $0.118$. The highest estimate was obtained by a shuffled-temporal control ($p_{\mathrm{Holm}}=0.057$), while the best non-randomized specification was the set-based graph model ($\rho=0.118$, $p_{\mathrm{Holm}}=0.090$), which uses neither observed order nor relational structure.
With sentence embeddings, estimates ranged from $0.101$ to $0.145$, and seven of the eight neural specifications yielded significant positive correlations after correction.
The observed-order GRU achieved the highest correlation ($\rho=0.145$, 95\% CI $[0.074,0.212]$, $p_{\mathrm{Holm}}=0.006$), compared with $0.128$ for its order-invariant counterpart and $0.100$ for the shuffled control.
For \TaSFC and \COH, no neural-model correlation was significant with either feature source, and no estimate exceeded $0.067$.

These correlations quantify absolute predictive performance rather than incremental contributions attributable to observed consultation order or relational structure.
No matched graph or sequence contrast was statistically significant after correction for either feature source.
The smallest Holm-adjusted $p$-value was $0.155$, for a contrast in which the temporal graph model performed worse than its set-based control ($\Delta\rho=-0.033$).
The largest individual contrast was observed for \COH with sentence embeddings, where the observed-order GRU outperformed its order-invariant counterpart by $\Delta\rho=0.048$ ($p=0.087$). However, this difference was not statistically significant even before multiple-testing correction.
For \DIRQ, the control models matched or exceeded the predictive performance of their observed-order counterparts in several cases.
The order-invariant GNN reached $\rho=0.135$, compared with $0.101$ for the observed-order temporal graph model. Conversely, the shuffled GRU reached only $0.100$, compared with $0.145$ for the observed-order GRU. The fixed-vector permutation control ($\mathcal{C}^\mathsf{C}+\mathcal{V}^\mathsf{C}+\widetilde{\mathcal{O}}^\mathsf{C}$) reached $0.126$, compared with $0.108$ for the corresponding observed-order specification.
These matched comparisons therefore provide no statistically significant evidence that observed consultation order or relational structure contributes additional predictive information about \DIRQ.
Across trace representations and estimators, Tier~3 predictive correlations for \DIRQ were heterogeneous rather than uniformly close to zero, ranging from $-0.079$ to $0.150$.
With Coh-Metrix features, fixed-vector estimators yielded correlations from $-0.079$ to $0.084$, compared with $0.077$ to $0.118$ for the neural models.
With sentence embeddings, random forests and support-vector regression generally yielded correlations between $0.09$ and $0.15$, whereas partial least squares remained near zero.
The prespecified Coh-Metrix--random-forest analysis therefore lies at the lower end of the observed performance range.

Several exploratory comparisons between model architectures were statistically significant after correction, primarily those involving the weakest fixed-vector estimator.
At Tier~3 in page set $\mathsf{C}$, the temporal graph model outperformed partial least squares on \DIRQ (95\% CI for the difference $[0.051,0.215]$, $p_{\mathrm{Holm}}=0.033)$.
The relational graph model also outperformed partial least squares (95\% CI $[0.043,0.228]$, $p_{\mathrm{Holm}}=0.040)$.
Neither corresponding comparison with random forest was significant after correction ($p_{\mathrm{Holm}}=0.090$ and $0.120$, respectively).
With sentence embeddings, the observed-order GRU outperformed partial least squares on \DIRQ ($p_{\mathrm{Holm}}=0.033$) and \TaSFC ($p_{\mathrm{Holm}}=0.050$).
As discussed in \Cref{sec:rq3}, these comparisons vary both model architecture and the information available to the model.
They therefore serve as descriptive architecture benchmarks rather than tests of the incremental contribution of observed consultation order or relational structure.

\begin{table}[htbp]
\centering
\caption{Composite construction. \DIRQ is the unweighted mean of \code{g2\_resp}, \code{g4\_2\_sourcesref} and \code{g5\_coherence}, rescaled by their codebook maxima (2 and 3; \code{g4\_2} by its highest value within the task). \TaSFC is the unweighted mean of the task's factual-correctness criteria. A missing value counts as zero only for bonus items, which are rated only when their parent criterion is fulfilled; sum items are dropped because they duplicate their parts.}
\label{tab:item-mapping}
\small
\begin{tabular}{lrrl}
\toprule
Task & \TaSFC criteria & of which bonus items & Excluded \\
\midrule
    FM & 9 & 3 & -- \\
    GS & 6 & 2 & -- \\
    HS & 7 & 0 & -- \\
    TP & 5 & 0 & \code{cq4\_1}, \code{cq4\_2} \\
\bottomrule
\end{tabular}
\end{table}

\paragraph*{\Rfinal{The largest correlations obtained.}}

\begin{table}[t]
    \centering
    \caption{Distribution of within-task Spearman correlations across all evaluated model configurations, grouped by predictor source and validation tier.}
\label{tab:ceiling}
    \small
    \begin{tabular}{@{}lrrrrr@{}}
        \toprule
        Specification & $k$ & Minimum & Mean & 95th percentile & Maximum \\
        \midrule
        All specifications & 6{,}192 & $-0.204$ & $0.063$ & $0.267$ & $0.453$ \\
        \addlinespace[0.5em]
        \multicolumn{6}{@{}l}{\textit{By predictor source, all tiers}} \\
        \quad Written response included & 1{,}512 & $-0.074$ & $0.180$ & $0.327$ & $0.453$ \\
        \quad\quad Response alone ($\mathcal{W}$) & 756 & $-0.064$ & $0.192$ & $0.327$ & $0.453$ \\
        \quad\quad Response with documents ($\mathcal{C}^\mathsf{C}+\mathcal{W}$) & 756 & $-0.074$ & $0.168$ & $0.332$ & $0.392$ \\
        \quad Browsing trace-only specifications & 4{,}680 & $-0.204$ & $0.026$ & $0.152$ & $0.246$ \\
        \addlinespace[0.5em]
        \multicolumn{6}{@{}l}{\textit{At Tier~3}} \\
        \quad All specifications & 2{,}064 & $-0.204$ & $0.056$ & $0.253$ & $0.364$ \\
        \quad Response alone ($\mathcal{W}$) & 252 & $-0.017$ & $0.174$ & $0.321$ & $0.363$ \\
        \quad Response with documents ($\mathcal{C}^\mathsf{C}+\mathcal{W}$) & 252 & $-0.064$ & $0.149$ & $0.289$ & $0.364$ \\
        \quad All trace-only specifications & 1{,}560 & $-0.204$ & $0.022$ & $0.137$ & $0.223$ \\
        \quad\quad Observed order or relations & 522 & $-0.198$ & $0.026$ & $0.142$ & $0.209$ \\
        \quad\quad Permutation or randomization controls & 522 & $-0.193$ & $0.027$ & $0.145$ & $0.223$ \\
        \bottomrule
    \end{tabular}
    \vspace{3pt}
    \begin{minipage}{0.96\linewidth}
        \footnotesize
        \textit{Note.} \(k\) denotes the number of model-configuration-level within-task correlation estimates, each averaged over the applicable evaluation repeats and random seeds.
        Trace-only specifications use predictors constructed solely from the browsing record.
        All estimates are out-of-fold.
        Rows pool both feature sources, the three page sets, all applicable estimators, and all outcomes.
        The last two rows summarize the observed-order and observed-relation specifications and their matched controls descriptively;
        inferential comparisons between them use the paired contrasts of \Cref{tab:conjunction-components}.
    \end{minipage}
\end{table}

\begin{table}[ht]
    \centering
    \caption{
    Within-task Spearman correlations for consulted-page content ($\mathcal{C}^\mathsf{C}$), written-response features ($\mathcal{W}$), and their concatenation ($\mathcal{C}^\mathsf{C}+\mathcal{W}$), across both feature sources and all three validation tiers.
    Results are based on random forests in page set $\mathsf{C}$ (514 scored student-task observations from 417 students).
    Brackets indicate 95\% student-cluster bootstrap confidence intervals; $p$-values are Holm-adjusted within the RQ1 absolute-correlation family.
    The written-response representations serve as exploratory baselines, whereas only the content-only correlation at Tier~3 belongs to the prespecified primary tests of absolute predictive performance.}
    \label{tab:response-spearman}
    \scriptsize
    \setlength{\tabcolsep}{3pt}
    \begin{tabular}{ll*{3}{lr}}
        \toprule
         & & \multicolumn{2}{c}{Tier 1, observation} & \multicolumn{2}{c}{Tier 2, student holdout} & \multicolumn{2}{c}{Tier 3, student and task holdout} \\
        \cmidrule(lr){3-4} \cmidrule(lr){5-6} \cmidrule(lr){7-8}
        Outcome & Representation & $\rho$ [95\% CI] & $p$ & $\rho$ [95\% CI] & $p$ & $\rho$ [95\% CI] & $p$ \\
        \midrule
        \multicolumn{8}{l}{\emph{Coh-Metrix indices}} \\
            \DIRQ & $\mathcal{C}^\mathsf{C}$ & $-0.029$ [-0.11, 0.05] & 1.000 & $-0.046$ [-0.13, 0.04] & 1.000 & $-0.079$ [-0.17, 0.01] & 1.000 \\
             & $\mathcal{W}$ & $+0.231$ [0.14, 0.31] & 0.006 & $+0.228$ [0.14, 0.31] & 0.006 & $+0.320$ [0.23, 0.40] & 0.006 \\
             & $\mathcal{C}^\mathsf{C}+\mathcal{W}$ & $+0.175$ [0.09, 0.26] & 0.006 & $+0.170$ [0.08, 0.25] & 0.006 & $+0.277$ [0.18, 0.36] & 0.006 \\
        \addlinespace[2pt]
            \TaSFC & $\mathcal{C}^\mathsf{C}$ & $+0.015$ [-0.07, 0.10] & 0.466 & $+0.032$ [-0.05, 0.12] & 0.513 & $-0.027$ [-0.12, 0.07] & 1.000 \\
             & $\mathcal{W}$ & $+0.122$ [0.04, 0.21] & 0.095 & $+0.120$ [0.03, 0.21] & 0.106 & $+0.126$ [0.03, 0.22] & 0.115 \\
             & $\mathcal{C}^\mathsf{C}+\mathcal{W}$ & $+0.102$ [0.02, 0.18] & 0.172 & $+0.112$ [0.03, 0.19] & 0.110 & $+0.077$ [-0.02, 0.16] & 1.000 \\
        \addlinespace[2pt]
            \COH & $\mathcal{C}^\mathsf{C}$ & $-0.047$ [-0.12, 0.03] & 1.000 & $-0.050$ [-0.13, 0.03] & 1.000 & $-0.039$ [-0.13, 0.05] & 1.000 \\
             & $\mathcal{W}$ & $+0.314$ [0.22, 0.39] & 0.006 & $+0.309$ [0.22, 0.39] & 0.006 & $+0.310$ [0.22, 0.39] & 0.006 \\
             & $\mathcal{C}^\mathsf{C}+\mathcal{W}$ & $+0.278$ [0.19, 0.36] & 0.006 & $+0.268$ [0.19, 0.35] & 0.006 & $+0.291$ [0.20, 0.38] & 0.006 \\
        \midrule
        \multicolumn{8}{l}{\emph{Sentence embeddings}} \\
            \DIRQ & $\mathcal{C}^\mathsf{C}$ & $+0.073$ [-0.01, 0.16] & 0.613 & $+0.058$ [-0.03, 0.14] & 1.000 & $+0.102$ [0.01, 0.18] & 0.115 \\
             & $\mathcal{W}$ & $+0.189$ [0.10, 0.27] & 0.006 & $+0.182$ [0.10, 0.27] & 0.006 & $+0.174$ [0.08, 0.27] & 0.006 \\
             & $\mathcal{C}^\mathsf{C}+\mathcal{W}$ & $+0.164$ [0.08, 0.24] & 0.006 & $+0.155$ [0.07, 0.24] & 0.006 & $+0.182$ [0.09, 0.27] & 0.006 \\
        \addlinespace[2pt]
            \TaSFC & $\mathcal{C}^\mathsf{C}$ & $+0.189$ [0.10, 0.27] & 0.006 & $+0.159$ [0.08, 0.24] & 0.006 & $-0.014$ [-0.10, 0.08] & 1.000 \\
             & $\mathcal{W}$ & $+0.296$ [0.22, 0.37] & 0.006 & $+0.290$ [0.22, 0.36] & 0.006 & $+0.031$ [-0.06, 0.12] & 1.000 \\
             & $\mathcal{C}^\mathsf{C}+\mathcal{W}$ & $+0.312$ [0.24, 0.38] & 0.006 & $+0.290$ [0.22, 0.36] & 0.006 & $+0.099$ [0.01, 0.19] & 0.476 \\
        \addlinespace[2pt]
            \COH & $\mathcal{C}^\mathsf{C}$ & $+0.074$ [-0.01, 0.16] & 0.935 & $+0.055$ [-0.03, 0.14] & 1.000 & $+0.008$ [-0.09, 0.10] & 1.000 \\
             & $\mathcal{W}$ & $+0.173$ [0.10, 0.25] & 0.006 & $+0.160$ [0.08, 0.24] & 0.006 & $+0.058$ [-0.03, 0.15] & 1.000 \\
             & $\mathcal{C}^\mathsf{C}+\mathcal{W}$ & $+0.144$ [0.06, 0.22] & 0.024 & $+0.139$ [0.05, 0.22] & 0.012 & $+0.069$ [-0.02, 0.16] & 1.000 \\
        \bottomrule
    \end{tabular}
\end{table}

\Rfinal{The main text focuses} on prespecified contrasts.
\Rfinal{This paragraph} summarizes the largest correlations observed across the complete analysis grid, providing an overview of predictive performance beyond the individual comparisons.
Across all 6{,}192 within-task correlations, spanning both feature sources, three page sets, three validation tiers, five estimators, and all outcomes, the maximum was $0.453$. This occurred when written-response features predicted source-reference use at Tier~1 in page set $\mathsf{S}$, using sentence embeddings and support-vector regression.
All correlations exceeding $0.4$ involved written-response features.
No trace-only representation exceeded $0.246$ across the analysis grid or $0.223$ at Tier~3.
Matched controls reached comparable values.
Among the 522 Tier~3 estimates in each group, the maximum was $0.209$ for observed-structure representations and $0.223$ for their controls. Their means ($0.026$ and $0.027$) and 95th percentiles ($0.142$ and $0.145$) were also similar.
The largest trace-only correlations occurred for task-standardized \TaSFC with sentence embeddings: $0.246$ for observed order at Tier~1 and $0.227$ for the consulted-page content representation at Tier~2. For the same specifications, the corresponding Tier~3 correlations were $-0.019$ and $0.003$, respectively.

\subsection*{\Rfinal{Criterion audit: further results}}

For the rater-sensitivity analysis only, criterion scores were recomputed twice: once using all raters and once excluding an unusually severe \COH rater. No correction for rater severity was applied.

The composites are not rated directly. Instead, raters score individual criteria, from which composite scores are calculated separately for each rater according to \Cref{sec:targets}, yielding one composite score per response and rater.
The variance decomposition is based on these rater-specific composite scores.
Cross-task consistency is estimated using data from the 136 students with valid scores on both tasks.
Each score is first converted to its rank among all valid responses to the same task, thereby removing task-specific differences in location and scale.
Each student then contributes one pair of ranks, with the two tasks ordered alphabetically rather than by completion time. A single Spearman correlation $\rho$ is computed across these pairs, pooling students from all six possible task combinations.
A two-sided test assesses the null hypothesis of zero Spearman correlation, with confidence intervals obtained using an approximate Fisher transformation and student-level bootstrapping.
For comparison, Spearman's $\rho$ is also computed directly from the original scores, since retaining task-specific differences in location and spread can reverse the sign of the pooled correlation.

\DIRQ averages three criteria whose cross-task consistencies differ (\Cref{fig:criterion-cross-task}).
\Rfinal{Source referencing} showed a correlation of $\rho=0.37$ (95\% CI $[0.21,0.51]$), followed by \COH ($\rho=0.30$); response fulfillment showed a weaker correlation ($\rho=0.12$, 95\% CI $[-0.05,0.27]$).
These criterion-level analyses were corrected separately from the confirmatory family.
\DIRQ's moderate cross-task consistency aligns more closely with source referencing and \COH than with response fulfillment, suggesting stronger contributions from these criteria to its task-general component.

We estimate the variance components using ratings from all 644 valid responses.
However, distinguishing between student-level variance and student-by-task interaction variance relies primarily on the 136 students with valid scores on two tasks, as their repeated observations provide information about cross-task consistency.
Observations from students who completed only one task inform the combined variation attributable to these components but provide limited information for separating them.
Consequently, the allocation of variance between the student and student-by-task components is particularly uncertain.
Rater-related variance components are estimated using ratings from the full sample.

Rater effects are the least precisely estimated components \Rfinal{in \Cref{tab:criterion-variance}}.
Six raters contributed substantive ratings, four providing 88\%; \Rfinal{one rated \COH unusually severely.}
A seventh, test account contributed one rating but remained a separate rater level.
Rater main effects are significant for \COH (19\%, $p=0.001$), not \TaSFC (2\%, $p=0.14$) or \DIRQ (9\%, $p=0.11$).
Their 95\% confidence intervals extend to 87\%, 16\%, and 60\%, respectively, indicating substantial uncertainty.
Rater-by-task interactions are significant for all outcomes ($p<0.001$), with narrower intervals \Rfinal{such as} [2\%, 14\%] for \TaSFC \Rfinal{and} [3\%, 29\%] for \DIRQ.
Rater variation appears limited for \TaSFC (2\%) and source-reference use (1\%), although confidence intervals extend to 16\% and 21\%, respectively.
For \COH, the rater component is larger (19\%) and statistically significant, but remains highly uncertain.
The rater main effect exceeds the rater-by-task interaction (19\% vs. 3\%), suggesting stable differences in rater severity; for response fulfillment, the reverse pattern (0\% vs. 15\%) suggests task-dependent differences.
Overall, \COH scores show greater estimated rater-related variance than \TaSFC or source-reference use.

The generalizability coefficient expresses the ratio of stable between-student variance to expected observed-score variance and, under the assumptions of Generalizability Theory, corresponds to the expected squared correlation between a student's observed score and their \Rfinal{universe score, averaged over all admissible tasks} \cite{Cronbach:et:al:1972}.
For a single task rated by three raters, the most common configuration in our sample, the estimated coefficients are $0.39$ for \DIRQ, $0.40$ for \COH, and $0.10$ for \TaSFC.
These estimates characterize the outcomes as operationalized in \Cref{sec:targets}, rather than the rubric itself: \COH is an individual rating criterion, whereas \DIRQ and \TaSFC are composites constructed from multiple criteria.
\TaSFC averages between 5--9 binary items, depending on the task, so differences in measurement length may contribute to the estimated student-by-task variance rather than reflecting task-dependent performance alone.
The remaining coefficients are projections for measurement designs not observed in this study: averaging scores across four tasks per student would yield an estimated generalizability coefficient of $0.70$ for \DIRQ and \COH, whereas \TaSFC would require approximately 22 tasks to reach the same level.
Under the fitted variance-component model, the main constraint on generalizability is therefore the number of tasks completed per student, rather than the number of raters per task.
All these coefficients presuppose a measurement model in which a task-general student standing can be defined and its generalizability estimated.
For \DIRQ and \COH, the stable between-student component is statistically distinguishable from zero, supporting the prospect of increasingly reliable task-general rankings as more tasks are sampled.
For \TaSFC, no such component is established; if its true value is zero, averaging over additional tasks cannot produce a reliable task-general ranking, regardless of how many tasks are administered.
The single-task generalizability coefficients largely reflect the cross-task consistency patterns already reported \Rfinal{in \Cref{sec:results-criterion-audit}} rather than providing independent evidence.
The projections may also be optimistic if person-by-rater variation is absorbed into the residual and persists across tasks, since such variation would not diminish through task averaging.

A post-hoc analysis of response length further qualifies the interpretation of \COH scores.
Whereas criterion-level analyses use all 644 validly rated responses, analyses requiring text-derived features are restricted to the 514 scored student-task observations in $\mathsf{C}$.
Within tasks, response length was associated with ratings, although association strength varied across criteria.
Averaged across the four tasks, Spearman's $\rho$ was $0.34$ for \COH, $0.27$ for source-reference use, and $0.11$ for response fulfillment, compared with a median of $0.03$ across the 32 factual-correctness criteria (\Cref{tab:response-length}).
The association was strongest among shorter responses.
Across within-task response-length quartiles, mean standardized \COH increased from $-0.66$ in the shortest quartile to $+0.37$ in the third, before declining slightly to $+0.27$ in the longest.
The corresponding standard deviation decreased from $1.22$ to $0.68$, then increased slightly to $0.76$.
The slightly lower mean and greater variability in the longest quartile are consistent with the possibility that greater response length does not necessarily imply greater coherence.
However, neither difference was statistically established: the mean difference was $-0.10$ (95\% CI $[-0.26,0.06]$), and the standard deviation difference was $+0.08$ (95\% CI $[-0.03,0.19]$).
The clearest descriptive pattern is that the shortest responses received lower and more variable \COH scores.

The design cannot distinguish between two potentially overlapping explanations.
Response length may be associated with competence, as maintaining coherence across more material is demanding, but raters may also be influenced by fluency and response length independently of coherence.
The measured text properties did not explain the association: after controlling for the seventeen response-level Coh-Metrix indices, the partial correlations of response length with \COH and \DIRQ were $0.337$ and $0.393$, respectively, compared with unadjusted correlations of $0.338$ and $0.391$.
Within each task, response length, the outcome, and the seventeen indices were rank-transformed. Response length and the outcome were then separately residualized against the indices, and the resulting residuals were correlated.
Task-specific correlations were averaged with equal weight.
Two additional findings suggest that the association cannot be attributed solely to raters' perceptions of response length.
Rater disagreement about \COH was not appreciably associated with response length ($\rho=-0.05$), while the partial correlation between response length and factual correctness increased from $0.11$ to $0.16$ after adjustment for the text indices.
The latter pattern is consistent with longer responses covering more factual-correctness checklist items.

Associations were robust to score construction: omitting codebook-range scaling barely changed \DIRQ's length correlation ($0.391$ to $0.398$), while treating a missing bonus item as missing rather than zero left \TaSFC's unchanged ($0.103$).

\subsection*{\Rfinal{Inference details}}

Because some students completed two tasks, bootstrap resampling and paired permutation tests use students as sampling units to preserve within-student dependence.
By contrast, permutation tests for individual correlations shuffle ratings independently within tasks, breaking cross-task correspondence between observations from the same student.
We construct 95\% confidence intervals using student-cluster percentile bootstrapping with \numprint{2000} resamples \cite{Efron:et:al:1993}.
We obtain two-sided $p$-values for specification contrasts from \numprint{5000} paired permutations, swapping the predictions of the two specifications within students.
We compute correlations within task, repeat, and neural-seed blocks, applying each student's swap consistently across all blocks containing their observations.
For individual correlations, we use the same bootstrap procedure but obtain one-sided $p$-values by permuting ratings within tasks.

An intersection--union test rejects the global null only when both component null hypotheses are rejected; its $p$-value is therefore the maximum of the two valid component $p$-values \cite{Berger:1982}.
Our component tests are two-sided, so small $p$-values can arise from effects in either direction.
To enforce the directional alternative, we set the conjunction $p$-value to the larger component $p$-value if both estimated effects are positive, and to 1 otherwise.

The task-wise permutation scheme is our declared procedure for testing individual correlations.
Because it does not preserve cross-task dependence among observations from the same student, we additionally evaluate the three RQ2 correlation tests against a student-block permutation null and report both results.
Under this alternative scheme, students are treated as exchangeable blocks: their rating vectors are permuted only among students who completed the same set of tasks, preserving both task membership and within-student dependence.
This follows the multi-level block permutation framework of \citeA{Winkler:et:al:2015}, with students as whole-block exchangeable units within strata defined by their completed task sets.
The 417 students with scored observations form ten such strata; for students observed on only one task, the procedure coincides with the declared task-wise permutation scheme.
For $\rho(\mathcal{C}^\mathsf{C})$, the declared $p$-values of $0.959$, $0.718$, and $0.788$ for \DIRQ, \TaSFC, and \COH change to $0.966$, $0.709$, and $0.802$, respectively, under student-block permutation.
The maximum absolute difference is $0.014$, and both procedures yield the same inferential conclusions.

\Rfinal{The conjunction rule of \Cref{sec:inference}, which takes the larger component $p$-value when both estimates are positive and 1 otherwise,} is conservative relative to a conjunction test based on valid one-sided component $p$-values, provided the two-sided tests appropriately control both tails.

Although step-down procedures control the family-wise error rate regardless of family size, their statistical power may decrease as the number of tests increases \cite{Holm:1979}.
For large-scale model comparisons involving thousands of tests, \citeA{Gardner:et:al:2018} discuss Bayesian alternatives.

Failure to reject the null hypothesis does not establish the absence of an effect.
The upper confidence bound nevertheless indicates how large a positive effect remains compatible with the data.
For the six individual primary contrasts, point estimates are close to zero, and their confidence intervals constrain the magnitude of positive effects compatible with the observed data.
We use the upper endpoint of the two-sided 90\% confidence interval as a one-sided 95\% upper confidence bound, corresponding to a significance level of 5\%.
This differs from the two-sided 95\% confidence intervals reported elsewhere.
All six \Rfinal{RQ2} estimates lie within $0.08$ of zero.
The one-sided 95\% upper confidence bounds are $0.16$, $0.11$, and $0.11$ for the correlations between consulted-page content and \DIRQ, \TaSFC, and \COH, respectively, and $0.09$, $0.13$, and $0.05$ for the corresponding incremental contributions beyond written-response features.
Across the twelve component contrasts underlying the order and relation conjunction tests, the one-sided 95\% upper confidence bounds range from $0.05$ to $0.16$.
The largest upper bound concerns the \TaSFC comparison between observed order and the order-invariant baseline, for which the estimated improvement was $0.098$ and the confidence interval excluded zero.
The conjunction nevertheless failed because observed order did not outperform its permutation control, rather than because both component contrasts were negligible.
Under the assumed effect sizes and variability, a replication of this design would have approximately 80\% power to detect a correlation of $0.13$ or an incremental correlation of $0.08$.
These estimates apply to individual, unadjusted tests, not to the Holm-adjusted test families or conjunction tests.
None of these analyses was prespecified: the analysis plan included no equivalence tests, and recomputing confidence bounds from the previously established intervals verifies their numerical consistency without making the resulting bounds part of the prespecified analysis.

\subsection*{Student-stable variance in the predictors}

The criterion audit decomposes the variance of the rated outcomes. The same question can be put to the predictors: how much of a feature is a stable property of the student across the tasks they completed? Identification rests on the 106 students whose two sessions were both eligible in page set $\mathsf{C}$, so the observation-level value of each feature is decomposed into a student component and a residual that confounds the student-by-task interaction with error, each student-task cell holding one session. Fits use \code{statsmodels} REML (\Rfinal{restricted maximum likelihood, REML}) with profile-likelihood intervals, as in the criterion audit.

Because a model uses a whole block rather than one feature, and because the coordinates of a sentence embedding carry no interpretation and its basis is arbitrary, the block-level quantity is also reported: the largest between-student share attainable by a linear combination, which is invariant to rotation of the feature space. Estimated in sample this is meaningless, since some direction separates the students whenever the block is wide, so it is cross-validated --- the combination is estimated from one set of students and the share measured on a disjoint set --- and compared with a null in which student identity is permuted within task. The cross-validated block estimates rest on 103 of these students, since a block requires a complete vector in both sessions. The embedding block is reduced by principal components to the width of the Coh-Metrix block so that the two are compared at equal dimension. This analysis is post hoc; the declaration names no such test.

\begin{table}[ht]
\centering
\caption{Between-student share of feature variance in page set $\mathsf{C}$, over the students observed on two tasks. Per-feature values are REML estimates with 95\% profile-likelihood intervals over 106 students, and the feature count is the number that survived decomposition: of the 31 session descriptors built \Rfinal{in $\mathsf{C}$ (of up to 34)}, 17 carry enough variance to be decomposed, the rest being annotation shares that are almost never observed. Block values are cross-validated over 103 and tested against 10{,}000 permutations of student identity within task. \Rfinal{Cross-validated shares can be negative when the combination estimated on one set of students does not separate the others.} $\mathcal{W}$ describes the written response, $\mathcal{C}$ the consulted pages, and $\mathcal{V}$ the session. A share near zero means the feature describes the task rather than the student.}
\label{tab:trace-generalizability}
\small
\begin{tabular}{@{}llrrr@{}}
\toprule
Block & Description & Features & Median & Largest single feature \\
\midrule
\Rfinal{$\mathcal{W}$} & Written response, 17 Coh-Metrix indices & 17 & $.196$ & $.475$ \code{WRDPRO} [.30, .67] \\
\Rfinal{$\mathcal{C}$} & Consulted pages, same 17 indices & 17 & $.000$ & $.135$ \code{WRDPRO} [.00, .28] \\
\Rfinal{$\mathcal{V}$} & Session descriptors & 17 & $.000$ & $.265$ log unique URLs [.11, .39] \\
\midrule
\multicolumn{5}{@{}l}{\emph{Block-level, cross-validated}} \\
 & & Dimensions & Share & $p$ \\
\Rfinal{$\mathcal{C}$} & Coh-Metrix content & 17 & $-.154$ & $.838$ \\
\Rfinal{$\mathcal{V}$} & Session descriptors & 31 & $+.262$ & $.013$ \\
\Rfinal{$\mathcal{C}$} & Sentence embeddings, PCA to 17 & 384 & $+.156$ & $.086$ \\
\bottomrule
\end{tabular}
\end{table}

A post-hoc analysis assessed cross-session predictor stability for 103 students with two eligible sessions (\Cref{tab:trace-generalizability}), without prespecified testing or multiplicity correction.
Consulted-page content showed no significant cross-session stability with Coh-Metrix or embeddings, unlike visit descriptors (\Rfinal{share $0.262$}, uncorrected $p=0.013$), although the responsible descriptors remain unidentified.
These engagement-related descriptors \cite{Gasevic:et:al:2016,Arizmendi:et:al:2023} suggest that browsing behavior may be more stable across tasks than consulted content.
Stability does not imply predictive validity: stable features may not predict performance, while task-dependent features may transfer to unseen tasks.
Nevertheless, Coh-Metrix indices showed greater between-student stability in written responses than consulted pages (median variance shares: $0.196$ vs. $0.000$).
This suggests stronger task constraints on consulted content, although the post-hoc interpretation is limited to the features, students, and tasks examined.

\subsection*{\Rfinal{Page-set sensitivity}}

Page set $\mathsf{A}$ serves for a prespecified sensitivity analysis including all retained pages, while $\mathsf{S}$ supports an exploratory analysis of search behavior.
SERPs are analyzed separately because they list search results rather than coherent documents.
Consequently, cohesion indices derived from SERPs and content documents require different interpretations.
Observations in page sets $\mathsf{C}$ and $\mathsf{S}$ retain their manifest assignments, enabling paired comparisons on matched observations within each page set.
The fixed-vector analyses are repeated in page sets $\mathsf{A}$ and $\mathsf{S}$.
Contrasts are computed within page sets; results across page sets are reported but not compared statistically.
We construct the same hypothesis families separately for %
sets $\mathsf{A}$ and $\mathsf{S}$.

\begin{figure}[ht]
    \centering
    \includegraphics[width=\textwidth]{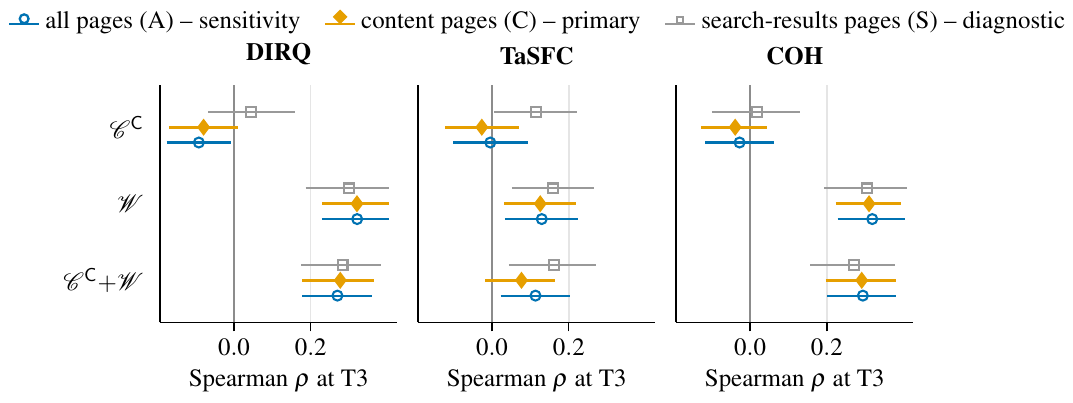}
    \caption{\Rfinal{RQ1--RQ2} at Tier~3 in the content-page \Rfinal{set $\mathsf{C}$} (primary), all-page \Rfinal{set $\mathsf{A}$} (sensitivity), and search-results \Rfinal{page set $\mathsf{S}$} (diagnostic). \Rfinal{Page sets} are presented side by side but are not tested against one another.}
    \label{fig:universes}
\end{figure}

The main findings were consistent across page sets (\Cref{fig:universes}).
In the full-process page set $\mathsf{A}$, which additionally includes search-engine results pages, the Tier~3 consulted-page content correlation ($\rho(\mathcal{C}^\mathsf{A})$) was $-0.091$ for \DIRQ, $-0.004$ for \TaSFC, and $-0.027$ for \COH.
The incremental contributions of consulted-page content beyond written-response features were again negative, and none of the conjunction tests was statistically significant.
In the search-results page set $\mathsf{S}$, the most notable result was the consulted-page content correlation ($\rho(\mathcal{C}^\mathsf{S})$) for \TaSFC ($0.113$, $p=0.025$), which was not significant after Holm correction ($p_{\mathrm{Holm}}=0.152$).
The corresponding conjunction tests were likewise not statistically significant.

One exploratory relational conjunction pattern survived multiple-testing correction in three feature-source and validation-tier configurations.
In the full-process page set $\mathsf{A}$, the relational conjunction for source-reference use was significant with support-vector regression at Tier~1 ($p_{\mathrm{Holm}}=0.025$) and Tier~2 ($p_{\mathrm{Holm}}=0.012$).
The corresponding Tier~1 test using the alternative feature source was also significant ($p_{\mathrm{Holm}}=0.044$).
These were the only significant relational conjunction results among 792 conjunction tests per feature source.
The pattern did not occur with either of the other estimators or in page sets $\mathsf{C}$ and $\mathsf{S}$, and it was absent at Tier~3, where evaluation requires transfer to a held-out task.
Given its exploratory status, limited robustness across configurations, and absence under task-transfer evaluation, we do not interpret this pattern as evidence that relational structure provides generalizable predictive information about source-reference use.

No estimator met the prespecified criterion for degeneracy.
Block-level correlations were defined for all 828 combinations of page set, feature source, validation tier, outcome, and estimator.
The rater-sensitivity analysis likewise left the main findings unchanged.
After reconstructing the outcomes without the unusually severe \COH rater, all conjunction tests remained non-significant, and the largest absolute change in a within-task correlation was $0.02$.
Remaining discrepancies between the reconstructed and supplied scores resulted from rebuilding the aggregates rather than specifically from excluding that rater.

\begin{table}[htbp]
\centering
\caption{Length of the analyzed page texts, by page class. A named source counts as opened when a page of the same host and path was opened in the same session, where one path may be a prefix of the other; the URLs are compared after removing the protocol, a leading \code{www.} or \code{m.}, the query string and a trailing slash, lower-casing host and path, and resolving search-engine click-through links to their target.}
\label{tab:text-lengths}
\small
\begin{tabular}{lrrrr}
\toprule
Page class & Visits & Median words & IQR & Median sentence length \\
\midrule
    Content pages & 3,383 & 759 & 486--1074 & 12.5 \\
    Search-results pages & 1,668 & 220 & 198--296 & 10.5 \\
\bottomrule
\end{tabular}
\end{table}

\begin{table}[htbp]
\centering
\caption{Cross-task rank consistency by task pair: Spearman correlation between a student's two scores, ranked within task, over the students who completed that pair. The pooled coefficients reported in the text are computed over all pairs jointly, not as an average of the coefficients in this table.}
\label{tab:task-pairs}
\small
\begin{tabular}{lrrrr}
\toprule
Task pair & $n$ & \DIRQ & \TaSFC & \COH \\
\midrule
    FM--GS & 23 & $+0.56$ & $+0.26$ & $+0.11$ \\
    FM--HS & 17 & $+0.57$ & $-0.27$ & $+0.42$ \\
    FM--TP & 23 & $+0.53$ & $-0.10$ & $+0.49$ \\
    GS--HS & 21 & $+0.29$ & $+0.17$ & $+0.31$ \\
    GS--TP & 27 & $+0.28$ & $+0.28$ & $+0.41$ \\
    HS--TP & 25 & $+0.07$ & $+0.18$ & $+0.18$ \\
\bottomrule
\end{tabular}
\end{table}

\begin{table}[htbp]
\centering
\caption{Variance components as percentages of the total variance of the composite scores (\Rfinal{restricted maximum likelihood, REML}), with the $p$-value of a likelihood-ratio test against the model without that component in parentheses. The test uses the one-sided boundary mixture of a point mass at zero and $\chi^2_1$; the residual has no test.}
\label{tab:variance-tests}
\small
\begin{tabular}{lrrrrr}
\toprule
Outcome & Student & Student $\times$ task & Rater & Rater $\times$ task & Residual \\
\midrule
    \DIRQ & 23 ($<$0.001) & 23 ($<$0.001) & 9 (0.112) & 8 ($<$0.001) & 37 \\
    \TaSFC & 8 (0.110) & 64 ($<$0.001) & 2 (0.139) & 5 ($<$0.001) & 21 \\
    \COH & 19 ($<$0.001) & 14 ($<$0.001) & 19 (0.001) & 3 ($<$0.001) & 45 \\
    Response fulfillment & 6 (0.049) & 20 ($<$0.001) & 0 (1.000) & 15 ($<$0.001) & 59 \\
    Source-reference use & 36 ($<$0.001) & 43 ($<$0.001) & 1 (0.022) & $<$1 ($<$0.001) & 20 \\
\bottomrule
\end{tabular}
\end{table}

\begin{table}[ht]
\centering
\caption{Post-hoc within-task Spearman correlations between written-response length (in words) and each rated criterion. The table gives the mean and range across the four tasks. Composite scores are shown under the declared construction and two counterfactual alternatives.}
\label{tab:response-length}
\scriptsize
\begin{tabular}{lrr}
\toprule
Rated criterion & $\rho$ & Range across tasks \\
\midrule
g5\_coherence & $+0.338$ & +0.25 to +0.50 \\
g4\_2\_sourcesref & $+0.270$ & +0.20 to +0.30 \\
g4\_sources & $+0.244$ & +0.19 to +0.27 \\
cq4\_values & $+0.166$ & +0.17 to +0.17 \\
cq1\_evidence & $+0.156$ & +0.16 to +0.16 \\
cq3\_sum & $+0.145$ & +0.14 to +0.14 \\
g3\_reco & $+0.137$ & +0.06 to +0.25 \\
cq7\_burns & $+0.135$ & +0.14 to +0.14 \\
cq3\_rad & $+0.135$ & +0.13 to +0.13 \\
g3\_sum & $+0.124$ & +0.07 to +0.23 \\
g2\_resp & $+0.113$ & +0.03 to +0.21 \\
cq6\_defic & $+0.102$ & +0.10 to +0.10 \\
cq3\_1 & $+0.098$ & +0.10 to +0.10 \\
cq3\_2 & $+0.092$ & +0.09 to +0.09 \\
cq2\_notconcl & $+0.080$ & +0.08 to +0.08 \\
cq2\_healtheffects & $+0.078$ & +0.08 to +0.08 \\
cq1\_recyclable & $+0.076$ & +0.08 to +0.08 \\
cq4\_composite & $+0.074$ & +0.07 to +0.07 \\
cq5\_disposal & $+0.072$ & +0.07 to +0.07 \\
cq4\_immune & $+0.048$ & +0.05 to +0.05 \\
cq3\_proteins & $+0.045$ & +0.05 to +0.05 \\
cq3\_intake & $+0.032$ & +0.03 to +0.03 \\
cq2\_sum & $+0.027$ & +0.03 to +0.03 \\
cq1\_2 & $+0.023$ & +0.02 to +0.02 \\
cq1\_sum & $+0.023$ & +0.02 to +0.02 \\
cq1\_borage & $+0.023$ & +0.02 to +0.02 \\
cq2\_notfully & $+0.016$ & +0.02 to +0.02 \\
cq1\_negeff & $+0.012$ & +0.01 to +0.01 \\
cq4\_sum & $+0.010$ & -0.03 to +0.05 \\
g3\_1\_corrreco & $+0.001$ & -0.07 to +0.15 \\
cq2\_mechanism & $-0.004$ & -0.00 to -0.00 \\
cq5\_limits & $-0.014$ & -0.01 to -0.01 \\
cq3\_overesti & $-0.019$ & -0.02 to -0.02 \\
g4\_1\_sourcesnoref & $-0.021$ & -0.04 to -0.01 \\
cq4\_1 & $-0.023$ & -0.09 to +0.04 \\
cq2\_1 & $-0.023$ & -0.02 to -0.02 \\
cq4\_radothersource & $-0.033$ & -0.03 to -0.03 \\
cq5\_pain & $-0.069$ & -0.07 to -0.07 \\
cq4\_2 & $-0.082$ & -0.08 to -0.08 \\
cq6\_funding & $-0.143$ & -0.14 to -0.14 \\
ampel & $-0.195$ & -0.27 to -0.10 \\
\midrule
\multicolumn{3}{l}{\emph{Composite scores}} \\
\DIRQ, declared & $+0.391$ & +0.34 to +0.42 \\
\DIRQ, declared criteria without codebook scaling & $+0.398$ & +0.38 to +0.42 \\
\TaSFC, declared & $+0.103$ & +0.06 to +0.14 \\
\TaSFC, missing bonus left missing & $+0.103$ & +0.06 to +0.14 \\
\COH facet, declared & $+0.338$ & +0.25 to +0.50 \\
\bottomrule
\end{tabular}
\end{table}

\begin{table}[ht]
\centering
\caption{Outcomes as scored and task-standardized in \Rfinal{page set $\mathsf{C}$}, using Coh-Metrix features and Random Forest. The upper panel gives within-task Spearman correlations at Tier~3. The lower panel gives \Rfinal{the pooling gap} at Tier~1, defined as the difference between pooled and within-task correlations calculated from the same predictions. The three outcomes are those with task-standardized counterparts --- \DIRQ, \TaSFC, and the Overall composite; the coherence rating has none and is not shown.}
\label{tab:task-standardized-results}
\small
\begin{tabular}{lrrrrrr}
\toprule
 & \multicolumn{2}{c}{\DIRQ} & \multicolumn{2}{c}{\TaSFC} & \multicolumn{2}{c}{Overall} \\
\cmidrule(lr){2-3}\cmidrule(lr){4-5}\cmidrule(lr){6-7}
Representation & as scored & standardized & as scored & standardized & as scored & standardized \\
\midrule
\multicolumn{7}{l}{\emph{Within-task $\rho$ at Tier~3}} \\
    \Rfinal{$\mathcal{C}$} & $-$0.079 & $-$0.149 & $-$0.027 & $-$0.077 & $-$0.013 & $-$0.097 \\
    \Rfinal{$\mathcal{W}$} & 0.320 & 0.327 & 0.126 & 0.127 & 0.195 & 0.193 \\
    \Rfinal{$\mathcal{C}+\mathcal{V}$} & 0.019 & 0.002 & $-$0.010 & $-$0.081 & $-$0.026 & $-$0.072 \\
    \Rfinal{$\mathcal{C}+\mathcal{V}+\mathcal{O}+\mathcal{G}$} & 0.004 & 0.021 & 0.055 & 0.021 & 0.077 & 0.043 \\
\midrule
\multicolumn{7}{l}{\emph{\Rfinal{Pooling gap} at Tier~1}} \\
    \Rfinal{$\mathcal{C}$} & 0.154 & 0.001 & 0.400 & 0.024 & 0.427 & 0.011 \\
    \Rfinal{$\mathcal{W}$} & 0.059 & 0.014 & 0.132 & 0.017 & 0.145 & 0.026 \\
    \Rfinal{$\mathcal{C}+\mathcal{V}$} & 0.147 & $-$0.008 & 0.422 & 0.017 & 0.446 & 0.008 \\
    \Rfinal{$\mathcal{C}+\mathcal{V}+\mathcal{O}+\mathcal{G}$} & 0.136 & $-$0.010 & 0.389 & 0.022 & 0.414 & 0.013 \\
\bottomrule
\end{tabular}
\end{table}

\begin{table}[ht]
\centering
\caption{The seventeen Coh-Metrix indices of the primary feature set, with the descriptions of the implementation used here. All are rates, incidences, or means, not raw counts. Incidences are counts per 1{,}000 words. \code{WRDFRQmc} is the implementation's Wikipedia-sample alternative to the CELEX-based frequency index. The same indices describe the written responses.}
\label{tab:coh-indices}
\small
\begin{tabular}{lll}
\toprule
Group & Index & Description \\
\midrule
    Connectives & \code{CNCADC} & Adversative and contrastive connectives \\
     & \code{CNCCaus} & Causal connectives \\
     & \code{CNCNeg} & Negative connectives \\
     & \code{CNCPos} & Positive connectives \\
    Descriptive & \code{DESWLlt} & Word length, number of letters, mean \\
    Syntactic Pattern Density & \code{DRAP} & Adverbial phrase density \\
     & \code{DRNEG} & Negation density \\
     & \code{DRNP} & Noun phrase density \\
     & \code{DRPP} & Preposition phrase density \\
     & \code{DRPVAL} & Agentless passive voice density \\
    Situation Model & \code{SMCAUSvp} & Causal verbs and causal particles \\
    Word Information & \code{WRDADJ} & Adjectives \\
     & \code{WRDADV} & Adverbs \\
     & \code{WRDFRQmc} & Wikipedia Log minimum frequency for content words, mean \\
     & \code{WRDNOUN} & Nouns \\
     & \code{WRDPRO} & Pronouns \\
     & \code{WRDPRP3p} & Third-person plural pronouns \\
\bottomrule
\end{tabular}
\end{table}

\clearpage
\printbibliography[title={References for the Supplementary Material}]

@techreport{Facione:1990,
	author      = {Facione, Peter A.},
	title       = {Critical Thinking: A Statement of Expert Consensus
	for Purposes of Educational Assessment and Instruction. {R}esearch Findings and Recommendations},
	year        = {1990},
	institution = {American Philosophical Association},
	type        = {The Delphi Report},
	number      = {ED315423},
	url         = {https://eric.ed.gov/?id=ED315423}
}

@article{Katz:2007,
	title={Testing information literacy in digital environments: {ETS}'s {iSkills} assessment},
	author={Katz, Irvin R},
	journal={Information Technology and Libraries},
	volume={26},
	number={3},
	pages={3--12},
	year={2007},
  doi = {10.6017/ital.v26i3.3271}
}

@book{VanDamme:Zahner:2022,
  editor    = {Van Damme, Dirk and Zahner, Doris},
  title     = {Does Higher Education Teach Students to Think Critically?},
  year      = {2022},
  publisher = {OECD Publishing},
  address   = {Paris},
  doi       = {10.1787/cc9fa6aa-en},
  url       = {https://doi.org/10.1787/cc9fa6aa-en}
}

@article{Tenison:Sparks:2023,
	author  = {Tenison, Caitlin and Sparks, Jesse R.},
	title   = {Combining cognitive theory and data driven approaches to examine students' search behaviors in simulated digital environments},
	journal = {Large-scale Assessments in Education},
	year    = {2023},
	volume  = {11},
	pages   = {28},
	doi     = {10.1186/s40536-023-00164-w},
	url     = {https://doi.org/10.1186/s40536-023-00164-w}
}

@article{Sparks:et:al:2016,
	author = {Sparks, Jesse R. and Katz, Irvin R. and Beile, Penny M.},
	title = {Assessing Digital Information Literacy in Higher Education: A Review of Existing Frameworks and Assessments With Recommendations for Next-Generation Assessment},
	journal = {ETS Research Report Series},
	volume = {2016},
	number = {2},
	pages = {1-33},
	doi = {10.1002/ets2.12118},
	url = {https://onlinelibrary.wiley.com/doi/abs/10.1002/ets2.12118},
	eprint = {https://onlinelibrary.wiley.com/doi/pdf/10.1002/ets2.12118},
	year = {2016}
}

@article{Shavelson:et:al:1993,
  author  = {Shavelson, Richard J. and Baxter, Gail P. and Gao, Xiaohong},
  title   = {Sampling Variability of Performance Assessments},
  journal = {Journal of Educational Measurement},
  volume  = {30},
  number  = {3},
  pages   = {215--232},
  year    = {1993},
  doi     = {10.1111/j.1745-3984.1993.tb00424.x}
}

@inproceedings{Errica:et:al:2020,
  author    = {Errica, Federico and Podda, Marco and Bacciu, Davide and Micheli, Alessio},
  title     = {A Fair Comparison of Graph Neural Networks for Graph Classification},
  booktitle = {Proceedings of the 8th International Conference on Learning Representations (ICLR)},
  year      = {2020},
  url       = {https://openreview.net/forum?id=HygDF6NFPB}
}

@book{McNamara:et:al:2014,
  author    = {McNamara, Danielle S. and Graesser, Arthur C. and McCarthy, Philip M.
               and Cai, Zhiqiang},
  title     = {Automated Evaluation of Text and Discourse with {Coh-Metrix}},
  publisher = {Cambridge University Press},
  year      = {2014},
  doi = {10.1017/CBO9780511894664}
}

@article{Graesser:et:al:2004,
  author  = {Graesser, Arthur C. and McNamara, Danielle S. and Louwerse, Max M.
             and Cai, Zhiqiang},
  title   = {{Coh-Metrix}: Analysis of text on cohesion and language},
  journal = {Behavior Research Methods, Instruments, \& Computers},
  volume  = {36},
  number  = {2},
  pages   = {193--202},
  year    = {2004},
  doi     = {10.3758/BF03195564}
}

@inproceedings{Reimers:et:al:2019,
  author    = {Reimers, Nils and Gurevych, Iryna},
  title     = {Sentence-{BERT}: Sentence Embeddings using Siamese {BERT}-Networks},
  booktitle = {Proceedings of the 2019 {Conference on Empirical Methods in Natural Language Processing} ({EMNLP-IJCNLP})},
  pages     = {3982--3992},
  year      = {2019},
  doi       = {10.18653/v1/D19-1410}
}

@inproceedings{Reimers:et:al:2020,
  author    = {Reimers, Nils and Gurevych, Iryna},
  title     = {Making Monolingual Sentence Embeddings Multilingual using
               Knowledge Distillation},
  booktitle = {Proceedings of the 2020 {Conference on Empirical Methods in Natural Language Processing} ({EMNLP})},
  pages     = {4512--4525},
  year      = {2020},
  doi       = {10.18653/v1/2020.emnlp-main.365}
}

@article{Roberts:et:al:2017,
	author = {Roberts, David R. and Bahn, Volker and Ciuti, Simone and Boyce, Mark S. and Elith, Jane and Guillera-Arroita, Gurutzeta and Hauenstein, Severin and Lahoz-Monfort, José J. and Schröder, Boris and Thuiller, Wilfried and Warton, David I. and Wintle, Brendan A. and Hartig, Florian and Dormann, Carsten F.},
	title = {Cross-validation strategies for data with temporal, spatial, hierarchical, or phylogenetic structure},
	journal = {Ecography},
	volume = {40},
	number = {8},
	pages = {913-929},
	doi = {10.1111/ecog.02881},
	url = {https://nsojournals.onlinelibrary.wiley.com/doi/abs/10.1111/ecog.02881},
	eprint = {https://nsojournals.onlinelibrary.wiley.com/doi/pdf/10.1111/ecog.02881},
	year = {2017}
}

@InProceedings{Koh:et:al:2021,
  title =  {{WILDS}: A Benchmark of in-the-Wild Distribution Shifts},  
  author =       {Koh, Pang Wei and Sagawa, Shiori and Marklund, Henrik and Xie, Sang Michael and Zhang, Marvin and Balsubramani, Akshay and Hu, Weihua and Yasunaga, Michihiro and Phillips, Richard Lanas and Gao, Irena and Lee, Tony and David, Etienne and Stavness, Ian and Guo, Wei and Earnshaw, Berton and Haque, Imran and Beery, Sara M and Leskovec, Jure and Kundaje, Anshul and Pierson, Emma and Levine, Sergey and Finn, Chelsea and Liang, Percy},  
  booktitle =  {Proceedings of the 38th {International Conference on Machine Learning}},  
  pages =  {5637--5664},  
  year =  {2021},  
  editor =  {Meila, Marina and Zhang, Tong},  
  volume =  {139},  
  series =  {Proceedings of Machine Learning Research},  
  publisher =    {PMLR},  
  url =  {https://proceedings.mlr.press/v139/koh21a.html}}

@inproceedings{Cho:et:al:2014,
    title = "Learning Phrase Representations using {RNN} Encoder{--}Decoder for Statistical Machine Translation",
    author = {Cho, Kyunghyun  and
      van Merri{\"e}nboer, Bart  and
      Gulcehre, Caglar  and
      Bahdanau, Dzmitry  and
      Bougares, Fethi  and
      Schwenk, Holger  and
      Bengio, Yoshua},
    editor = "Moschitti, Alessandro  and
      Pang, Bo  and
      Daelemans, Walter",
    booktitle = "Proceedings of the 2014 {Conference on Empirical Methods in Natural Language Processing} ({EMNLP})",
    year = "2014",
    address = "Doha, Qatar",
    publisher = "Association for Computational Linguistics",
    url = "https://aclanthology.org/D14-1179/",
    doi = "10.3115/v1/D14-1179",
    pages = "1724--1734"
}

@inproceedings{Gilmer:et:al:2017,
author = {Gilmer, Justin and Schoenholz, Samuel S. and Riley, Patrick F. and Vinyals, Oriol and Dahl, George E.},
title = {Neural message passing for Quantum chemistry},
year = {2017},
publisher = {PMLR},
booktitle = {Proceedings of the 34th {International Conference on Machine Learning}},
volume = {70},
url = {https://proceedings.mlr.press/v70/gilmer17a.html},
pages = {1263–1272},
series = {Proceedings of Machine Learning Research}
}

@article{Breiman:2001,
  author  = {Breiman, Leo},
  title   = {Random Forests},
  journal = {Machine Learning},
  year    = {2001},
  volume  = {45},
  number  = {1},
  pages   = {5--32},
  doi     = {10.1023/A:1010933404324},
  url     = {https://doi.org/10.1023/A:1010933404324},
  issn    = {1573-0565}
}

@article{Holm:1979,
  author  = {Holm, Sture},
  title   = {A Simple Sequentially Rejective Multiple Test Procedure},
  journal = {Scandinavian Journal of Statistics},
  volume  = {6},
  number  = {2},
  pages   = {65--70},
  year    = {1979},
  url = {https://www.jstor.org/stable/4615733}
}

@book{Efron:et:al:1993,
  author    = {Efron, Bradley and Tibshirani, Robert J.},
  title     = {An Introduction to the Bootstrap},
  publisher = {Chapman \& Hall},
  year      = {1993},
  doi = {10.1007/978-1-4899-4541-9}
}

@article{Berger:1982,
  author  = {Berger, Roger L.},
  title   = {Multiparameter Hypothesis Testing and Acceptance Sampling},
  journal = {Technometrics},
  volume  = {24},
  number  = {4},
  pages   = {295--300},
  year    = {1982},
  doi     = {10.1080/00401706.1982.10487790}
}

@incollection{Rouet:et:al:2011,
  author    = {Rouet, Jean-Fran{\c c}ois and Britt, M. Anne},
  title     = {Relevance Processes in Multiple Document Comprehension},
  booktitle = {Text Relevance and Learning from Text},
  editor    = {McCrudden, Matthew T. and Magliano, Joseph P. and Schraw, Gregory},
  pages     = {19--52},
  publisher = {Information Age Publishing},
  address   = {Charlotte, NC},
  year      = {2011},
  doi = {10.1108/978-1-61735-531-820251003}
}

@article{Molerov:et:al:2020,
  author  = {Molerov, Dimitri and Zlatkin-Troitschanskaia, Olga and Nagel, Marie-Theres and Br{\"u}ckner, Sebastian and Schmidt, Susanne and Shavelson, Richard J.},
  title   = {Assessing University Students' Critical Online Reasoning Ability: A Conceptual and Assessment Framework With Preliminary Evidence},
  journal = {Frontiers in Education},
  volume  = {5},
  pages   = {577843},
  year    = {2020},
  doi     = {10.3389/feduc.2020.577843}
}

@article{Nagel:et:al:2022,
  author  = {Nagel, Marie-Theres and Zlatkin-Troitschanskaia, Olga and Fischer, Jennifer},
  title   = {Validation of newly developed tasks for the assessment of generic Critical Online Reasoning (COR) of university students and graduates},
  journal = {Frontiers in Education},
  volume  = {7},
  pages   = {914857},
  year    = {2022},
  doi     = {10.3389/feduc.2022.914857}
}

@article{Nagel:2025,
  author    = {Nagel, Marie-Theres and Zlatkin-Troitschanskaia, Olga},
  title     = {The competence to deal critically with online information among economics education students and teacher trainees in economics: implications for teacher training},
  journal   = {Empirical Research in Vocational Education and Training},
  year      = {2025},
  volume    = {17},
  number    = {1},
  pages     = {11},
  doi       = {10.1186/s40461-025-00186-4},
  url       = {https://doi.org/10.1186/s40461-025-00186-4},
  publisher = {Springer Nature}
}

@article{Hartig:et:al:2026,
  author  = {Hartig, Johannes and Drake, Philine and Illmann, Jannick and K{\"o}hler, Carmen and Kohmer, Anika and Stroganova, Alisa},
  title   = {Rating procedures to evaluate generic critical online reasoning in an open {I}nternet environment},
  journal = {Zeitschrift f{\"u}r Erziehungswissenschaft},
  volume  = {29},
  number  = {1},
  pages   = {19--44},
  year    = {2026},
  doi     = {10.1007/s11618-025-01365-5}
}

@article{List:et:al:2019,
  author  = {List, Alexandra and Alexander, Patricia A.},
  title   = {Toward an Integrated Framework of Multiple Text Use},
  journal = {Educational Psychologist},
  volume  = {54},
  number  = {1},
  pages   = {20--39},
  year    = {2019},
  doi     = {10.1080/00461520.2018.1505514}
}

@book{Brennan:2001,
  author    = {Brennan, Robert L.},
  title     = {Generalizability Theory},
  publisher = {Springer},
  address   = {New York},
  year      = {2001},
  doi = {10.1007/978-1-4757-3456-0}
}

@book{Cronbach:et:al:1972,
  title     = {The Dependability of Behavioral Measurements:
               Theory of Generalizability for Scores and Profiles},
  author    = {Cronbach, Lee J. and Gleser, Goldine C. and
               Nanda, Harinder and Rajaratnam, Nageswari},
  year      = {1972},
  address   = {New York},
  publisher = {John Wiley \& Sons},
  isbn      = {0471188506}
}

@article{Schmidt:et:al:2020,
  author  = {Schmidt, Susanne and Zlatkin-Troitschanskaia, Olga and Roeper, Jochen and Klose, Verena and Weber, Maruschka and B{\"u}ltmann, Ann-Kathrin and Br{\"u}ckner, Sebastian},
  title   = {Undergraduate Students' Critical Online Reasoning---Process Mining Analysis},
  journal = {Frontiers in Psychology},
  volume  = {11},
  pages   = {576273},
  year    = {2020},
  doi     = {10.3389/fpsyg.2020.576273}
}

@article{Wang:et:al:2023,
  author  = {Wang, Karen D. and Cock, Jade Ma{\"i} and K{\"a}ser, Tanja and Bumbacher, Engin},
  title   = {A Systematic Review of Empirical Studies Using Log Data from Open-Ended Learning Environments to Measure Science and Engineering Practices},
  journal = {British Journal of Educational Technology},
  volume  = {54},
  number  = {1},
  pages   = {192--221},
  year    = {2023},
  doi     = {10.1111/bjet.13289}
}

@article{Gardner:et:al:2018,
  author  = {Gardner, Joshua and Brooks, Christopher},
  title   = {Evaluating Predictive Models of Student Success: Closing the Methodological Gap},
  journal = {Journal of Learning Analytics},
  volume  = {5},
  number  = {2},
  pages   = {105--125},
  year    = {2018},
  doi     = {10.18608/jla.2018.52.7}
}

@article{Xu:et:al:2026,
  author  = {Xu, Alice and Zhang, Icy and Blake, Adam B. and Stigler, James W.},
  title   = {Modelability as a Strategy for Improving the Generalizability and Scalability of Predictive Models},
  journal = {Journal of Learning Analytics},
  volume  = {13},
  number  = {1},
  pages   = {89--109},
  year    = {2026},
  doi     = {10.18608/jla.2026.9099}
}

@article{Winkler:et:al:2015,
  author  = {Winkler, Anderson M. and Webster, Matthew A. and Vidaurre, Diego and Nichols, Thomas E. and Smith, Stephen M.},
  title   = {Multi-Level Block Permutation},
  journal = {NeuroImage},
  volume  = {123},
  pages   = {253--268},
  year    = {2015},
  doi     = {10.1016/j.neuroimage.2015.05.092}
}

@article{Yu:et:al:2025,
  author  = {Yu, Linyu and Halpin, Peter F. and Bernacki, Matthew L. and Ren, Sirui and Plumley, Robert D. and Greene, Jeffrey A.},
  title   = {Interpreting Predictive Learning Sequences in a College Math Course through a Self-Regulated Learning Framework},
  journal = {Journal of Learning Analytics},
  volume  = {12},
  number  = {3},
  pages   = {66--86},
  year    = {2025},
  doi     = {10.18608/jla.2025.8865}
}

@article{Gasevic:et:al:2016,
title = {Learning analytics should not promote one size fits all: The effects of instructional conditions in predicting academic success},
journal = {The Internet and Higher Education},
volume = {28},
pages = {68-84},
year = {2016},
issn = {1096-7516},
doi = {10.1016/j.iheduc.2015.10.002},
url = {https://www.sciencedirect.com/science/article/pii/S1096751615300038},
author  = {Ga{\v{s}}evi{\'c}, Dragan and Dawson, Shane and Rogers, Tim and Gasevic, Danijela}
}

@article{Arizmendi:et:al:2023,
  author  = {Arizmendi, Cara J. and Bernacki, Matthew L. and Rakovi{\'c}, Mladen and Plumley, Robert D. and Urban, Christopher J. and Panter, A. T. and Greene, Jeffrey A. and Gates, Kathleen M.},
  title   = {Predicting Student Outcomes Using Digital Logs of Learning Behaviors: Review, Current Standards, and Suggestions for Future Work},
  journal = {Behavior Research Methods},
  volume  = {55},
  number  = {6},
  pages   = {3026--3054},
  year    = {2023},
  doi     = {10.3758/s13428-022-01939-9}
}
\end{refsection}

\end{document}